\documentclass[sigconf, 10pt]{acmart}
\usepackage{hyperref}
\usepackage{graphicx}
\usepackage{caption}
\usepackage{adjustbox}
\usepackage{mathtools}  
\usepackage{amsmath}
\usepackage{paralist}
\usepackage{multirow}
\usepackage{graphicx}
\usepackage{subcaption}
\usepackage{float} 
\usepackage{bbding}
\usepackage{pifont}

\usepackage[ruled, linesnumbered, noend]{algorithm2e}

\newcommand{\sysname}[0]{{\sc V-Droid}\xspace}
\newcommand{\eg}{{\it e.g.,}\xspace}
\newcommand{\ie}{{\it i.e.,}\xspace}

\copyrightyear{2026}
\acmYear{2026}
\setcopyright{cc}
\setcctype{by}
\acmConference[MobiCom '26]{The 32nd Annual International Conference on Mobile Computing and Networking}{October 26--30, 2026}{Austin, TX, USA}
\acmBooktitle{The 32nd Annual International Conference on Mobile Computing and Networking (MobiCom '26), October 26--30, 2026, Austin, TX, USA}
\acmPrice{}
\acmDOI{10.1145/3795866.3796681}
\acmISBN{979-8-4007-2505-0/2026/10}

\begin{document}
\title{\sysname: Advancing Mobile GUI Agent Through Generative Verifiers}

\author{\normalfont Gaole Dai}
\authornote{Work done during internship at Microsoft Research.}
\affiliation{%
  \institution{Nanyang Technological University}
  \country{}
}
\email{gaole001@e.ntu.edu.sg}

\author{Shiqi Jiang}
\authornote{Corresponding authors.}
\affiliation{%
  \institution{Microsoft Research}
  \country{}
}
\email{shijiang@microsoft.com}

\author{Ting Cao}
\affiliation{%
  \institution{Tsinghua University}
  \country{}
}
\email{tingcao@air.tsinghua.edu.cn}

\author{Yuanchun Li}
\affiliation{%
  \institution{Tsinghua University}
  \country{}
}
\email{liyuanchun@air.tsinghua.edu.cn}

\author{Yuqing Yang}
\affiliation{%
  \institution{Microsoft Research}
  \country{}
}
\email{yuqyang@microsoft.com}

\author{Rui Tan}
\affiliation{%
  \institution{Nanyang Technological University}
  \country{}
}
\email{tanrui@ntu.edu.sg}

\author{Mo Li}
\authornotemark[2] %
\affiliation{%
  \institution{HKUST}
  \country{}
}
\email{lim@cse.ust.hk}

\author{Lili Qiu}
\affiliation{%
  \institution{Microsoft Research}
  \country{}
}
\email{liliqiu@microsoft.com}

\renewcommand{\shortauthors}{Gaole Dai, Shiqi Jiang, Ting Cao, Yuanchun Li et al.}
\renewcommand{\shorttitle}{\sysname: Advancing Mobile GUI Agent Through Generative Verifiers}



\begin{abstract}

    We propose \sysname, a mobile GUI task automation agent. Unlike previous mobile agents that utilize Large Language Models (LLMs) as generators to directly generate actions at each step, \sysname employs LLMs as verifiers to evaluate candidate actions before making final decisions. To realize this novel paradigm, we introduce a comprehensive framework for constructing verifier-driven mobile agents: the discretized action space construction coupled with the prefilling-only workflow to accelerate the verification process, the pair-wise progress preference training to significantly enhance the verifier's decision-making capabilities, and the scalable human-agent joint annotation scheme to efficiently collect the necessary data at scale. 
    
    \sysname obtains a substantial task success rate across several public mobile task automation benchmarks: 59.5\% on AndroidWorld, 38.3\% on AndroidLab, and 49\% on MobileAgentBench, surpassing existing agents by 5.2\%, 2.1\%, and 9\%, respectively. Furthermore, \sysname achieves a remarkably low latency of 4.3s per step and 0.7s per decision, which are $6.1\times$ and $32.1\times$ faster compared with existing mobile agents. 
    The source code is available at \url{https://github.com/V-Droid-Agent/V-Droid}.

\end{abstract}

\maketitle

\setcounter{table}{0}
\setcounter{figure}{0}

\section{Introduction}
\label{sec:intro}

Controlling mobile devices via natural language has been a longstanding aspiration in the mobile domain~\cite{wen2024autodroid, lee2024mobilegpt, wenAutoDroidV2BoostingSLMbased2024, shvoAppBuddyLearningAccomplish2021}, promising opportunities to automate repetitive tasks and elevate user convenience. 
Unlike API-based agents that rely on predefined function calls~\cite{li2023api,chen2024octopus}, mobile GUI agents simulate human interactions, allowing them to operate across diverse applications through Graphical User Interface (GUI). However, developing such an agent poses significant challenges: it needs to not only interpret on-screen content but also make reasonable decisions to execute multi-step tasks within dynamic and complex GUI environments.

\begin{figure}[!t]
    \centering
    \includegraphics[width=0.8\linewidth]{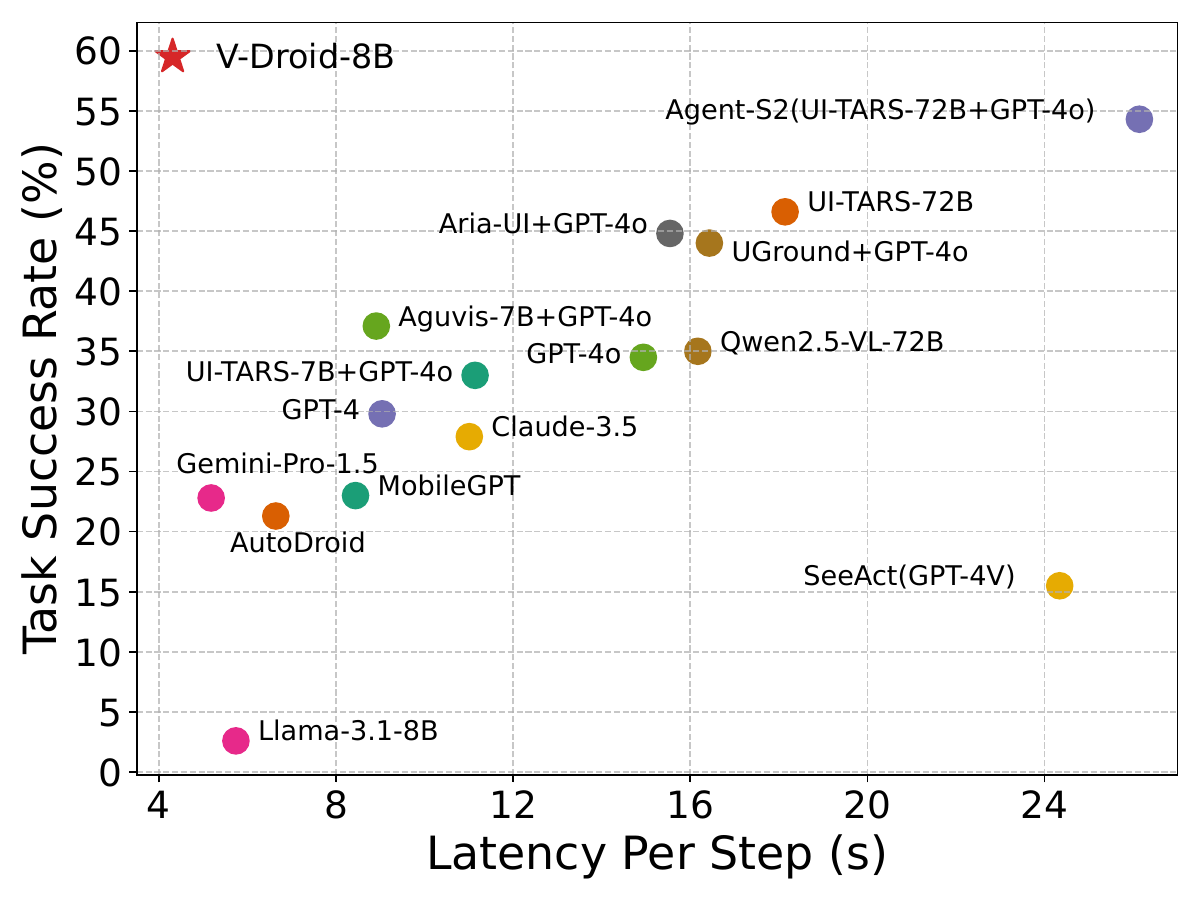}
    \caption{\textcolor{black}{Task success rate and latency per step of current mobile agents ~\protect\footnotemark and \sysname evaluated on AndroidWorld benchmark. The latency of 2B, 7B and 8B agents are measured on  $2\times$ Nvidia 4090. For 72B or MoE agents, the latency is measure on $4\times$ Nvidia A100 80G.}}
    \label{fig:performance_android_world}
\end{figure}

\textcolor{black}{In recent years, a variety of LLM-powered mobile GUI agents have been proposed~\cite{wen2024autodroid,agashe2025agents,qin2025ui,gou2025uground,zheng2024seeact,wang2024distrl,wang2024ponder,hong2024cogagent,yang2024aria}. These agents typically \emph{utilize LLMs as generators}, generating decisions (\eg reasoning and actions) based on the current task states (\eg user interfaces, task descriptions), leveraging the contextual understanding and reasoning abilities inherent to LLMs. These agents, however, fall short of meeting practical deployments. Fig.\ref{fig:performance_android_world} presents the performance of state-of-the-art (SOTA) mobile agents in terms of task success rate (SR) and stepwise latency on the AndroidWorld \cite{rawles2024androidworld}. As shown, the highest SR achieved by existing agents is only 54.3\%, significantly lower than the human performance of 80\%. Additionally, latency remains a critical challenge. Agent-S2~\cite{agashe2025agents}, powered by GPT-4o and UI-TARS-72B~\cite{qin2025ui}, takes over 25 seconds for a single step. }

\footnotetext{The task success rate achieved by GPT-4 and 4o, Claude, Llama, Gemini and Qwen is measured with the default prompt templates \ie T3A and M3A, in AndroidWorld benchmark.}

The suboptimal SR is predominantly hindered by the limited decision-making capabilities of existing agents when performing mobile tasks. While techniques such as prompt engineering~\cite{wen2024autodroid,wei2022chain} and GUI fine-tuning~\cite{gou2025uground,yang2024aria,qin2025ui} are commonly employed, these methods primarily focus on enhancing general performance of LLM, such as instruction-following, or refining agents' adaptability to mobile interfaces. However, they fail to adequately address task-specific, multi-step decision-making challenges, particularly in the context of mobile GUI control.

The high latency of existing mobile agents primarily stems from the autoregressive decoding mechanism of LLMs. These agents generate multiple tokens sequentially for each decision at every step. For instance, SeeAct~\cite{zheng2024seeact} generates around 100 tokens per step for one decision. Moreover, techniques like Chain-of-Thought (CoT) \cite{wei2022chain} and ReAct \cite{yao2023react} are widely employed to enhance reasoning and mitigate hallucination, further increasing output length, worsening latency issues.

\textcolor{black}{The fundamental research problem in developing practical mobile agents lies in enhancing their decision-making capabilities for mobile tasks while simultaneously maintain reasonable latency. To achieve this goal, we introduce \sysname, which, as illustrated in Fig.~\ref{fig:performance_android_world}, achieves a marked improvement in task success rate, refreshing the record to 59.5\% on the AndroidWorld benchmark, while achieves $6.1\times$ speed up on the step-wise latency.}

\begin{figure}[t]
    \centering
    \includegraphics[width=1.0\linewidth]{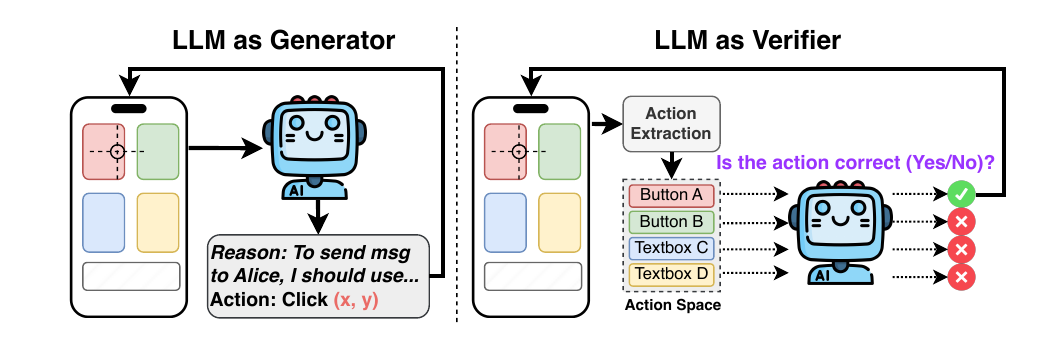}
    \caption{\textcolor{black}{The key differences in agent architecture between using LLMs as generators and as verifiers for decision-making: rather than directly determine actions based on states,  verifier-driven agents explicitly evaluate each action before arriving at the decision.}}
    \label{fig:generator_verifier}
\end{figure}

The key idea behind \sysname is \emph{transforming the paradigm of mobile agents by using LLM as verifiers instead of generators}, which is illustrated in Fig.~\ref{fig:generator_verifier}: rather than directly generating the final decision, when LLMs are used as verifiers, potential actions are first extracted. The verifier-driven agents then explicitly evaluate each candidate action \eg by prompting, "\emph{Is action X helpful for completing the task}?" Following these evaluations, the agent selects the action with the highest value, \eg the likelihood of generating a 'Yes' token.

Essentially, the verifier-driven architecture decouples one action decision-making process into two discrete processes: action extraction and action verification. This decoupling offers substantial advantages for advancing mobile agents:
\begin{inparaenum}[(i)]
    \item Intuitively, verifying an answer is much easier than generating one from scratch, a phenomenon known as the generation-verification gap~\cite{song2025mindgapexaminingselfimprovement, lightmanLetsVerifyStep2023}. Instead of directly making decisions within an expansive and infinite action space, the verifier-driven agent evaluates actions within an extractable and enumerable space, thereby simplifying the decision-making process. More importantly, for mobile devices, the interactive UI elements is rather limited (see \S~\ref{subsec:opportunity}) at each task step.
    \item The verification process typically requires generating fewer tokens, such as simple outputs like "Yes" or "No", which dramatically reduces latency. Moreover, actions to be verified can be processed in batches, fully leveraging hardware parallelism and further improving efficiency.
\end{inparaenum}

However, to fully unleash the potential of verifier-driven mobile agents, several technical challenges must be addressed: 
\begin{inparaenum}[(i)]
    \item Effectively extracting action from UI and constructing clean and complete action space, while efficiently verifying multiple available actions at each task execution step.
    \item Designing effective training methods for the verifier to improve decision-making capabilities, particularly since directly utilizing pre-trained LLMs as verifiers is inadequate as detailed in \S~\ref{subsec:opportunity}.
    \item Collecting and organizing the necessary data to support the training process at scale.
\end{inparaenum}

To address these challenges, \sysname introduces holistic designs for building a verifier-driven mobile agent, including:

\textbf{Verifier-driven agent workflow.} At each task step, the workflow of \sysname encompasses three main stages: extracting the action space, scoring with the verifier, and executing the selected action. First, we introduce a lightweight action extractor capable of accurately constructing and augmenting the action space for subsequent verification based on GUI representations, \eg the Android Accessibility Tree~\cite{Android_AccessibilityWindowInfo}. Next, available actions are verified using a prefilling-only approach, thus eliminating the constraints on LLM decoding. Moreover, multiple actions are verified in batches, leveraging prefix caching to significantly reduce latency. Finally, the action with the highest estimated score is executed.

\textbf{Pair-wise process preference ($P^3$) training method.} To improve the decision-making capabilities of \sysname, we introduce a pair-wise preference training strategy tailored for verifiers. Unlike prior post-training approaches that rely solely on mobile GUI data, our training leverages labeled mobile task trajectories with fine-grained process supervision~\cite{lightmanLetsVerifyStep2023}. It enables the verifier to prioritize correct actions by assigning them higher scores while penalizing incorrect actions at each task step. This approach significantly enhances the task-specific decision-making proficiency of \sysname.

\textbf{Scalable human-agent joint annotation scheme.} To collect the required fine-grained task trajectories, which is absent in existing datasets, we propose a human-agent joint annotation scheme based on the observation: when \sysname verifies a group of actions at each task step, the entropy of the assigned scores strongly correlates with step-wise correctness. Thus, we utilize a trained verifier to produce initial annotations, limiting human involvement to correcting only the erroneous annotations identified by the agent. Moreover, the annotation and training process is conducted iteratively, allowing the agent to be progressively trained on increasingly larger datasets, thereby effectively minimizing human labeling efforts while scaling the training process.

\textcolor{black}{The verifier in \sysname, which uses the small language model (SLM), Llama-3.1-8B, as the backbone, is trained using the $P^3$ method across four iterative training rounds with 110K samples collaboratively annotated by humans and agents. We evaluate \sysname on three public, realistic task benchmarks: AndroidWorld \cite{rawles2024androidworld}, AndroidLab \cite{xu2024androidlab}, and the MobileAgentBench~\cite{wang2024mobileagentbench}. On these benchmarks, \sysname refreshes the task success rates to 59.5\%, 38.3\%, and 49\%, surpassing previous best-performing agents by absolute margins of 5.2\%, 2.1\%, and 9.0\%, respectively. Compared to other SLM-based agents, \sysname achieves significant SR improvements of 50\%, 22\%, and 15\% on these benchmarks. In addition, \sysname delivers a 6.1$\times$ speedup over prior SOTA mobile agents. }

\textcolor{black}{In summary, we make the following contributions:}
\begin{itemize}
    \item We introduce \sysname, the first verifier-driven mobile agent framework, accompanied by comprehensive design principles.  
    \item We propose the pair-wise process preference training method, demonstrating its effectiveness in enhancing the decision-making capabilities for mobile GUI task.  
    \item We develop a human-agent joint annotation approach, enabling scalable training of mobile agents. 
    \item \textcolor{black}{\sysname significantly outperforms previous SOTA in task success rates on multiple public benchmarks while reducing latency by 6.1$\times$.  }
\end{itemize}

\section{Related Work and Motivation}
\label{sec:motivation}

\subsection{LLM Powered Mobile GUI Agent}
\label{subsec:background}

Recently, the emergence of numerous LLM-powered mobile agents~\cite{ wen2024autodroid, wang2024mobile, dingMobileAgentEnhancingMobile2024, wangMobileAgentBenchEfficientUserFriendly2024, li2024appagent, li2024appagent} has been observed. The introduction of large language models (LLMs) has significantly improved the ability of mobile agents to comprehend context and generate effective actions. Existing agents typically employ the following strategies to enhance performance.

\textcolor{black}{Most mobile agents~\cite{zheng2024seeact, wen2024autodroid, wang2024mobile, dingMobileAgentEnhancingMobile2024, wangMobileAgentBenchEfficientUserFriendly2024, li2024appagent, li2024appagent} leverage prompt engineering techniques to optimize task execution. For instance, SeeAct~\cite{zheng2024seeact} employs a ReAct-style~\cite{yao2023react} prompt to break down tasks into manageable steps, thereby reducing errors and mitigating hallucinations in the LLM’s outputs. Additionally, agents like AutoDroid~\cite{wen2024autodroid} and MobileGPT~\cite{lee2024mobilegpt} collect task completion traces during offline preprocessing stages for specific applications. These traces are integrated with the memory of pre-trained LLMs, enabling enhanced performance tailored to particular applications.
}

\textcolor{black}{
Several recent works, such as UGround~\cite{gou2025uground}, Aria-UI~\cite{yang2024aria}, UI-TARS~\cite{qin2025ui}, and Ferret-UI~\cite{youFerretUIGroundedMobile2024}, propose training grounding models that leverage pre-trained LLMs to comprehensively interpret and interact with the UI. These approaches exploit the inherent reasoning capabilities of LLMs for mobile tasks. However, despite the rapid advances in LLMs driven by large-scale pre-training, the lack of training corpora specifically tailored to mobile GUI tasks leaves existing models insufficiently prepared for reliable mobile task automation, as demonstrated in Fig.~\ref{fig:performance_android_world}.}

In addition to utilizing pre-trained LLMs for mobile agents, researchers have explored GUI fine-tuning strategies to improve task execution~\cite{liu2024autoglmautonomousfoundationagents, qwen2025qwen25technicalreport}. This involves adapting a pre-trained language model using domain-specific data, such as annotated screenshots and GUI representations. While GUI fine-tuning enhances the model's ability to accurately interpret and interact with GUI elements~\cite{zhangLlamaTouchFaithfulScalable2024, gao2024mobileviewslargescalemobilegui}, it remains insufficient for facilitating task-specific reasoning and multi-step decision-making, especially in dynamic and complex GUI environments, as evidenced in Fig.~\ref{fig:performance_android_world}.

Apart from the accuracy, few existing studies address latency optimization for mobile agents. MobileGPT~\cite{lee2024mobilegpt} attempts to cache execution traces for tasks successfully completed. However, this approach lacks scalability due to the diverse range of tasks and applications. As illustrated in Fig.~\ref{fig:performance_android_world}, most existing agents require more than 10 seconds per step, highlighting significant inefficiencies in task execution.

The fundamental challenges of improving decision-making capabilities while reducing decision-making latency for mobile agents remain unresolved.

\subsection{Opportunity and Challenges}
\label{subsec:opportunity}

\textcolor{black}{We tackle this challenge by proposing a novel approach: \emph{using LLMs as verifiers instead of generators} for mobile agents.}

\begin{figure}[t]
    \centering
    \includegraphics[width=0.8\linewidth]{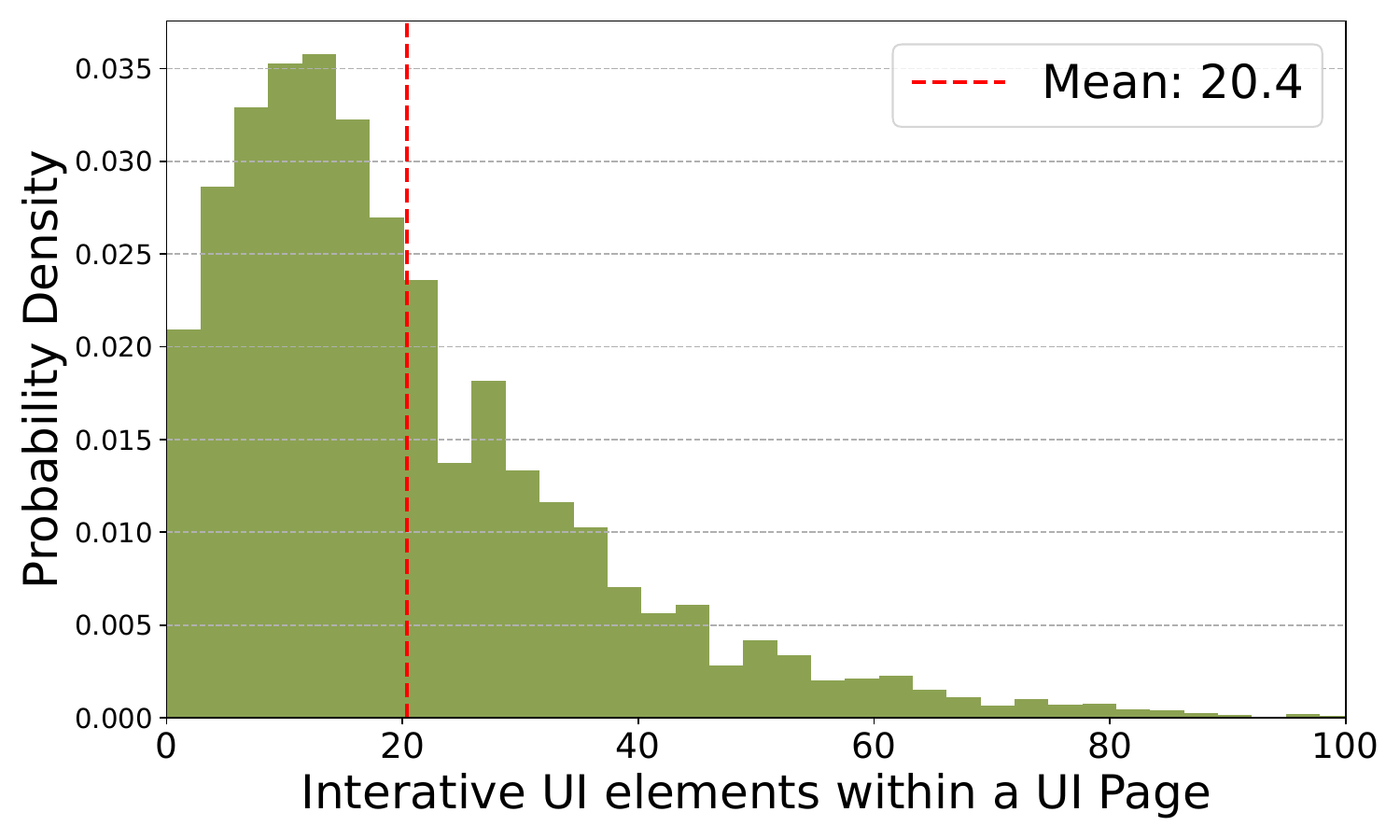}
    \caption{The distribution of interactive UI elements within each UI page by analyzing around $25,000$ real-world UI screens from the public dataset~\cite{li2024effects}.}
    \label{fig:action_distribution}
    \vspace{-1em}
\end{figure}

\textcolor{black}{Fig.~\ref{fig:generator_verifier} illustrates the key difference in agent architecture between using LLMs as generators and using LLMs as verifiers. The feasibility of verifier-driven agents depends on the presence of an enumerable and extractable action space. In the context of GUI automation tasks on mobile devices, this prerequisite is entirely met. Owing to the constrained screen size and the inherent interaction patterns of touchscreen interfaces, both the types of actions and the number of interactive UI elements on a single page are limited.}

\textcolor{black}{Fig.\ref{fig:action_distribution} illustrates the distribution of interactive UI elements within each GUI state. As depicted, the interactive action space on mobile devices is generally constrained to approximately $20$ elements on average. Although we have a unique opportunity to build verifier-driven agents for mobile tasks, several technical challenges are not well addressed.}

\textcolor{black}{Firstly, the detailed architecture and workflow of the verifier-driven agent remain ambiguous. In particular, constructing a well-defined action space is a nontrivial challenge. The supported action type varies on different UI elements and certain useful actions may not be explicitly visible, \eg \textit{navigate home}. Furthermore, some actions are inherently continuous rather than discrete, \eg \textit{type text}. Beyond that, efficiently verifying a set of actions also poses difficulties, especially considering the best utilizing the hardware parallelism.}

Secondly, directly utilizing a pre-trained LLM as the verifier is not a feasible solution for mobile agents. For instance, Llama-3.1-8B, fail to complete any tasks within the benchmark (detailed in \S~\ref{subsec: design alternatives}). Even powerful LLMs, \ie GPT-4 as the verifier, achieve only $34.5$\% task success rate on the AndroidWorld benchmark. The challenge of effectively fine-tuning such a verifier remains unresolved.

Thirdly, the data necessary to enable effective training, particularly fine-grained labeled task trajectories, is, to the best of our knowledge, absent in publicly available datasets. The collection and annotation of such data at scale present a significant challenge, posing a critical obstacle in the development of verifier-driven agents.

To this end, in \sysname, we introduce a comprehensive framework that integrates holistic designs to develop verifier-driven mobile agents, encompassing the agent architecture, training methodologies, and data collection strategies. In the following, we present these components in detail, starting with the workflow of \sysname.

\section{Workflow of \sysname}
\label{sec:architecture}

\begin{figure}
    \centering
    \includegraphics[width=1.0\linewidth]{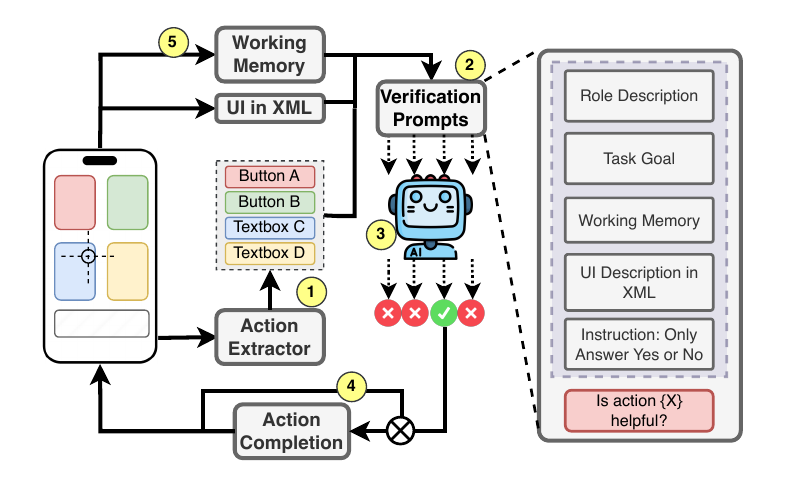}
    \caption{\textcolor{black}{The Workflow of \sysname: \ding{172} Extracting actions from UI and supplementing default actions; \ding{173} Constructing verification prompts with the template for candidate actions; \ding{174} Scoring with the verifier in batch with prefix caching; \ding{175} Completing and executing the selected action; \ding{176} Updating the working memory.}}
    \label{fig:details of verifier} 
\end{figure}

Fig.~\ref{fig:details of verifier} illustrates the workflow of \sysname. As a verifier-driven agent, \sysname requires the enumeration of candidate actions prior to estimating the optimal action at each step. To achieve this, \sysname employs a rule-based action candidate extractor to obtain actions from the current UI state. Each action candidate is then encapsulated using a predefined prompt template. Following this, the fine-tuned LLM as the verifier is utilized to evaluate and assign scores to these candidates in batch. The action with the highest score is selected and executed. In the following, we provide an in-depth explanation of each module.

\subsection{Constructing Action Space} 
A task $\kappa$ is automated through a sequence of steps. At any given step $t$, we extract a set of actions $\mathcal{A}_\kappa(t)$, representing the potential interactions available within the UI state $\mathcal{S}_\kappa(t)$ associated with the current step $t$, collectively defining the action space. This action space generally comprises two categories: UI-dependent and -independent actions, the latter being default actions that are not explicitly visible.

\textcolor{black}{
The UI-dependent action space is defined by the basic action types and the interactive UI elements available in the current UI. Owing to the inherent nature of touchscreen interactions on mobile devices, the set of basic action types is relatively limited. Specifically, we identify the following basic actions: \textit{click}, \textit{long-press}, \textit{scroll}, \textit{type text}, and \textit{clear text}. 
}

\textcolor{black}{To identify interactive UI elements, the XML representation of the UI state $\mathcal{S}_\kappa(t)$ is directly analyzed, extracted using the Android Accessibility Service~\cite{Android_AccessibilityWindowInfo}. A rule-based extractor is applied to detect UI elements, such as button, checkbox, and textbox, that are clickable, long-clickable, scrollable, or editable. These identified elements are then mapped to their corresponding basic action types. For example, and editable text box $\beta$ would be mapped to the action \textit{`input \{content\} to $\beta$'}. To prevent a combinatorial explosion of candidates, the standard on-screen keyboard is disabled, and text input is instead performed directly via Android commands~\cite{Android_Command}.}

\textcolor{black}{In addition to the UI-dependent actions, there exist actions that are not visible in the current UI but are essential for device control and task completion. Specifically, we supplement every action space $\mathcal{A}_\kappa(t)$ with the following default actions: \textit{open apps}, \textit{wait}, \textit{navigate home}, \textit{navigate back}, \textit{complete task}, and \textit{answer users' question}.}

\textcolor{black}{The analysis presented in \S~\ref{sec:discussion} demonstrates that the rule-based extraction employed by \sysname effectively encompasses the majority of interactions observed in real-world mobile tasks, thereby ensuring comprehensive practical coverage without unnecessarily complicating the action space.}

\subsection{Scoring with Verifier}

Once the action space $\mathcal{A}_\kappa(t)$ is determined, each action $\alpha\in\mathcal{A}_\kappa(t)$ is formatted into a predefined prompt template $P$ to construct the corresponding verification prompt $\rho_\kappa^t(\alpha)$. The structure of the prompt template is illustrated in Fig. \ref{fig:details of verifier} and includes the following key components of descriptions: role, goal, working memory, UI state, instruction, and the specific question to be verified.

In the role component, we outline the guidelines for operating mobile devices, as in~\cite{rawles2024androidworld}. The goal component specifies the task to be automated. The working memory maintains a record of action histories and trajectories for the current task, and it is updated after the execution of each action. For the UI component, we provide a streamlined description of the UI in XML format, as in ~\cite{wen2024autodroid}. The instruction component delivers explicit directives for the generative verifier, specifying that it must respond strictly with 'Yes' or 'No'. Finally, the prompt concludes with the specific question: \textit{Is the action $\alpha$ helpful for completing the given user task?}

Subsequently, all the formatted verification prompts, each corresponding to an action candidate $\alpha$ within the action space $\mathcal{A}_\kappa(t)$, are fed in batch to the verifier $\mathcal{V}$. The verifier is fine-tuned from a pre-trained LLM, \ie Llama-3.1-8B. The fine-tuning process enables the verifier to assign a score to each action by analyzing only the first generated token, \eg the possibility of 'Yes' or 'No'. Ideally, this score reflects the likelihood of successfully completing the task if the corresponding action were executed at the current step $t$. The training details of the verifier are elaborated in Section \ref{sec:training}.

Finally, after evaluating action candidates, the action $\alpha_\kappa^y(t)$ assigned the highest score is selected for execution at $t$:
\begin{equation}
    \alpha_\kappa^y(t) =\arg\max\mathcal{V}(\rho_\kappa^t(\alpha)), \alpha\in\mathcal{A}_\kappa(t)
\end{equation}

\subsection{Accelerating Verifications}

Assigning scores to a single action requires only one token generated by the generative verifier, such the pre-filling-only architecture significantly enhances the efficiency of mobile agents.  When verifying a set of actions at each step, all verifications can be processed in parallel as a batch. Moreover, it is observed in Fig.~\ref{fig:details of verifier} that, at a given task step, nearly all components of the verification prompts, apart from the question, remain identical. This design aims to maximize the shared prefix in the prompt, which can be leveraged to further accelerate the inference process.

Prefix caching enables the reuse of key-value (KV) caches across multiple verifications, thereby eliminating the need to recompute costly intermediate results. We adopt the Automatic Prefix Caching (APC) in vLLM~\cite{kwon2023efficientmemorymanagementlarge}. Building on prefix caching, we further group actions to optimize the verification process. Specifically, within each batch of actions, we first perform a warm-up verification. Subsequently, actions of the same type are grouped together for verification. Since verifications of identical action types share a greater portion of their prompt content, this approach increases cache utilization and further reduces latency. 

\subsection{Completing and Executing Action}
Before the selected action is exactly executed, three specific types of actions, \textit{open app}, \textit{type text}, and \textit{answer}, require additional completion to specify the target app to open, the content to type, and the response to the user's query. Thus we employ an LLM to generate the necessary content by prompting the current UI state and working memory.

Action completion introduces extra overhead to the prefill-only agent architecture in \sysname. But the number of actions requiring completion is relatively small. From the measurement on real world $2,000$ tasks in \cite{li2024effects}, only 12.4\% actions required completion. Furthermore, the separation of action selection and completion in \sysname significantly simplifies the action space, enhancing the overall efficiency.

The selected and completed actions are executed by simulating interactions such as clicking, long-clicking, scrolling, and other corresponding operations. This execution results in a new UI state, $\mathcal{S}(t+1)$. Following the reflection framework proposed in ~\cite{shinn2023reflexion}, a step-level summary is generated by recording the executed action. The working memory is then updated with this summary. Subsequently, \sysname iteratively performs the extracting actions, scoring with the verifier,  and executing workflow until the agent signals task completion by reporting the \textit{complete task} action.

\section{Pairwise Process Preference Training}
\label{sec:training}

\textcolor{black}{
The generative verifier in \sysname, which assigns a score to each action, enables the agent to discern which actions are more likely to contribute effectively to completing the given task. A straightforward approach involves directly employing a pre-trained LLM as the verifier without any post-training, using the logits or the probability of the 'Yes' token from the output space as the verification score. However, our measurements (\S\ref{subsec: design alternatives}) indicate that directly leveraging the Llama-3.1-8B as the verifier results in a success rate as low as 0\% on AndroidWorld benchmark \cite{rawles2024androidworld}.}

\textcolor{black}{
The rationale behind the suboptimal performance is twofold. First, for the verifier without post-training, we observe that the scores derived from the token space are insufficiently distinguishable, particularly in scenarios with numerous action candidates. Furthermore, simple fine-tuning on GUI data fails to enhance this indistinguishability. For instance, Qwen2.5-VL-72B~\cite{qwen2025qwen25technicalreport} fine-tuned with Android GUI data~\cite{rawlesAndroidWildLargeScale2023} achieves only 35\% success rate. Therefore, we believe that, to bridge the gap between the generated token space and the desired scoring (reward) space, additional training is necessary. More importantly, this training must be task-specific rather than relying solely on domain adaptation with GUI data, as the decision-making capabilities of mobile agent cannot be effectively enhanced otherwise. We propose the Pairwise Process Preference ($P^3$) training to significantly enhance the decision-making capabilities of verifier-driven mobile agents.
}

\subsection{Decision-Making Training}
\label{sec: p3_training}

\begin{figure}
    \centering
    \includegraphics[width=1.0\linewidth]{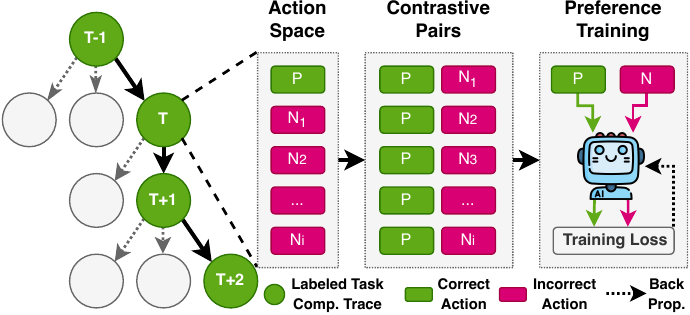}
    \caption{Illustration of $P^3$ training used in \sysname.}
    \label{fig:p3_training_illustration}
\end{figure}

The objective of $P^3$ training is to maximize the distinction between the action to be selected and the actions to be not selected at each step, guided by the process supervision~\cite{lightmanLetsVerifyStep2023}. To achieve this, training samples are structured into positive-negative action pairs at each step. By utilizing these contrastive pairs, $P^3$ training empowers the verifier to learn to assign higher scores to positive actions and lower scores to negative ones within the given context. Next, we delve into the training details, starting with the format of training data.

\textbf{Training data format}. Fig.~\ref{fig:p3_training_illustration} illustrates the process of organizing the training data. For a given task $\kappa$, $P^3$ training necessitates fine-grained process labels, \ie the trace for completing task $\kappa$ with the labeled correct action at each step. Specifically, the action space $\mathcal{A}_\kappa(t)$ is derived from at each step $t$, as described in Section \ref{sec:architecture}. Within the action space $\mathcal{A}_\kappa(t)$, the labeled action is identified as the positive action $\alpha_\kappa^y(t)$, while the remaining actions constitute the set of negative samples $\mathcal{A}^\prime_\kappa(t)$. Subsequently, contrastive training pairs $\Gamma(t)$ are constructed for step $t$ as:
\begin{equation}
    \Gamma_\kappa(t)=\{(\rho_\kappa^t(\alpha_\kappa^y(t)), \rho_\kappa^t(\alpha_\kappa^n(t)))\;|\;\alpha_\kappa^n(t)\in\mathcal{A}^\prime_\kappa(t)\},
\end{equation}
where $\rho_\kappa^t$ represents the prompts constructed at step $t$ using the prompt template. These prompts incorporate the positive action $\alpha_s^t$ and the negative action $\alpha_j^t$, respectively, along with the UI state $\mathcal{S}_\kappa(t)$, task description, working memory, and other contextual information, as detailed in Section \ref{sec:architecture}. Finally, we aggregate all the training pairs generated at every step of task $\kappa$ across all labeled tasks to construct the complete training dataset.

It is worth noting that, to the best of our knowledge, no publicly available dataset contains the required fine-grained process labels with contrastive action pairs. Therefore, we propose a novel data collection and synthesis approach, which will be detailed in~\S \ref{section:data_engineering}.

\textbf{Training loss design}.
The following loss is adopted to train the generative verifier at each step in \sysname:
\begin{equation} 
\label{equ: p3_loss}
    L_\kappa(t) = - E_{(\alpha_\kappa^y(t), \alpha_\kappa^n(t))}[\log\sigma(\tau(\alpha_\kappa^y(t))-\tau(\alpha_\kappa^n(t)))],
\end{equation}
\textcolor{black}{where $\alpha_\kappa^n(t)\in\mathcal{A}^\prime_\kappa(t)$ and $\tau$ represents the assigned score for the verified action. Specifically, in \sysname, a trainable MLP layer is appended to the LLM to project its output token probabilities into the estimated score for each action.}
\begin{equation}
    \tau(\alpha) = MLP\cdot\mathcal{V}(\rho_\alpha).
\end{equation}

\textcolor{black}{When the $P^3$ loss $L_\kappa(t)$ at each step $t$ of task $\kappa$ across all labeled tasks is minimized during training, the verifier learns to maximize the score assigned to the positive action while minimizing the score assigned to the negative action. }

\textcolor{black}{In P$^3$ training, we introduce three key concepts: \emph{pairwise}, \emph{process}, and \emph{preference}, all of which are crucial for effectively training the verifier. Firstly, preference training offers greater discriminative ability for the verifier to distinguish similar UI or actions which can be otherwise misleading to mobile agents. Secondly, process supervision provides fine-grained supervision signals, which are inherently beneficial for improving the step-wise decision-making ability of the verifier. Furthermore, compared to outcome-based supervision~\cite{lightmanLetsVerifyStep2023}, process supervision allows for the construction of a larger number of training samples. For instance, from 100 task traces, we can derive around 55K step-wise process labels. Thirdly, pairwise training can be considered a form of data augmentation. Without such augmentation, only one training sample might be obtained for each task step. In contrast, pairwise training enables the generation of 50$\times$ more training samples, which is especially critical when the available training data is limited.}

\begin{figure}
    \centering
    \includegraphics[width=1.0\linewidth]{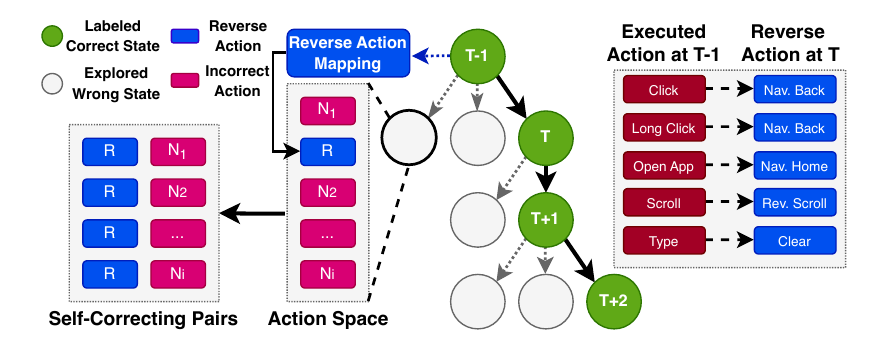}
    \caption{\textcolor{black}{Illustration of constructing self-correcting pairs and the mapping relationship between executed action and reverse action.}}
    \label{fig:self-correcting training}
\end{figure}

\subsection{Self-Correcting Training}
\label{sec: self-correct training}

In $P^3$ training, only the action pairs corresponding to the states on the correct trace are utilized. This raises an intriguing question: can we also harness the data from the states on incorrect traces, as illustrated in Fig.~\ref{fig:self-correcting training}.

Such data are valuable in enhancing the \emph{self-correcting} capability of mobile agents. Intuitively, when humans interact with devices, it is possible to enter an erroneous state unrelated to the current intent. Instead of abandoning the task and starting over, humans often try to correct such mistakes, \eg by clicking the \emph{navigate back} button. To the best of our knowledge, no existing mobile agent possesses the self-correcting capability. But the action pairs derived from these erroneous states present an opportunity for us to equip \sysname with self-correcting ability through $P^3$ training. 

Specifically, as illustrated in Fig.~\ref{fig:self-correcting training}, given that the state $\mathcal{S}_{t-1}$ at step $t-1$ lies on the correct trace, the state $\mathcal{S}_t$ can be reached by executing the labeled correct action $\alpha_\kappa^y(t-1)$. Conversely, executing any incorrect action $\alpha_\kappa^n(t-1) \in \mathcal{A}_\kappa^\prime(t-1)$ would lead to an erroneous state for the given task $\kappa$, noted as $\mathcal{S}^\prime(t)$.

On the erroneous state $\mathcal{S}^\prime(t)$, none of the actions from the action space may be applicable. Instead, a \emph{reverse action} $\alpha_\kappa^r(t)$ is introduced as the correct action, \eg pressing the navigate back button. Specifically, the reverse action is derived based on the executed action at the previous step $t-1$ through a predefined mapping relationship, as illustrated in Fig.~\ref{fig:self-correcting training}. For example, when \sysname enters incorrect content by executing the action \textit{type text}, the corresponding reverse action is to clear the previously input content.

Based on the identified reverse actions, we further construct the contrastive training pairs on the erroneous states for self-correcting training as follows:
\begin{equation}
    \Gamma^\prime_\kappa(t)=\{(\rho_\kappa^t(\alpha_\kappa^r(t)), \rho_\kappa^t(\alpha_\kappa(t)))\;|\;\alpha_\kappa(t)\in\mathcal{A}_\kappa(t)\}.
\end{equation}
The $\alpha_\kappa^r(t)$ represents the derived reverse action,
\begin{equation}
    \alpha_\kappa^r(t) = \mathcal{M}(\alpha_\kappa^y(t-1)),
\end{equation}
where $\mathcal{M}$ represents the mapping function shown in Fig.~\ref{fig:self-correcting training}. Ultimately, the self-correcting training pairs are combined with the contrastive training pairs obtained in Section~\ref{sec: p3_training} to form the final training set for \sysname.

It is worth noting that the number of erroneous states exceeds the number of correct states for a given task. We find training with excessive self-correcting pairs would lead to collapsed performance by continuing output the \textit{navigating back} action. Therefore, we randomly sample the erroneous states to construct the self-correcting training pairs, which account for approximately 2.5\% of the entire training set.

\section{Human-Agent-Joint-Annotation}
\label{section:data_engineering}

\begin{figure}[t]
    \centering
    \includegraphics[width=1.0\linewidth]{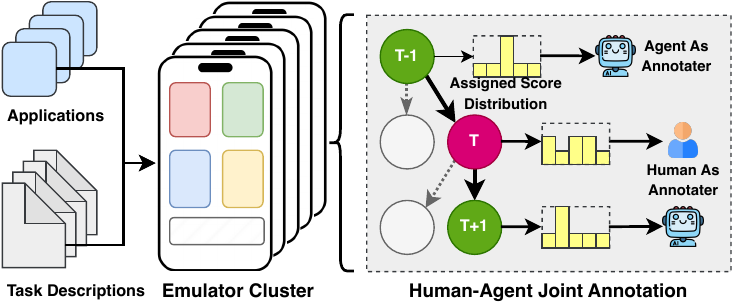}
    \caption{Illustration of human-agent joint annotation.}
    \label{fig:human_agent_anno}
\end{figure}

\textcolor{black}{$P^3$ training enhances the capabilities of mobile agents. However, to the best of our knowledge, the necessary training data, particularly fine-grained process labels that capture both correct and erroneous states, remains unavailable in public repositories. To this end, we propose a novel data synthesis methodology, as illusrated in Fig.~\ref{fig:human_agent_anno}. First, we collect task instructions for Android Apps from public datasets~\cite{li2024effects, rawlesAndroidWildLargeScale2023}. Additionally, we leverage LLMs to synthesize additional task instructions based on the Apps' descriptions available on the App Store, followed by meticulous manual verification. Subsequently, we generate the training data by actually executing the targeted tasks using the corresponding applications on our cluster of Android emulators. Human and agents jointly annotate the correct actions required at each step towards successful task completion.}

The key idea of the human-agent joint annotation is to leverage the trained \sysname to perform initial annotations, with human involvement only to correcting incorrect annotations, as illustrated in Fig.~\ref{fig:human_agent_anno}. Furthermore, we divide the annotation and training process into multiple iterations, allowing the agent to be progressively trained with larger datasets, thereby reducing its annotation errors over time. This iterative approach enables us to constrain human labeling efforts while effectively scaling the training process.

\subsection{Verifier-As-Annotator}
\label{subsec: human-in-loop}

To facilitate effective agent-human collaborative annotation, the critical challenge is determining when and at which specific step the agent is likely to produce incorrect annotations, thereby necessitating human intervention.

We observe that when our trained verifier evaluates a set of actions at a given step, there exists a strong correlation between the ambiguity of the score distribution and the step-wise correctness of the agent. Intuitively, when the verifier assigns a single high score to one action while assigning significantly lower scores to the others, the action with the highest score is likely correct. However, if the score distribution is inconsistent, with multiple actions receiving similar scores, the decision at this step is more likely to be incorrect.

Particularly, we employ the entropy to measure the ambiguity of the score distribution $\mathcal{E}$ at the step $t$,
\begin{equation}
    \mathcal{E}_\kappa(t) = -\sum\tau(\alpha_\kappa(t))\log(\tau(\alpha_\kappa(t))) | \alpha_\kappa(t)\in\mathcal{A}_\kappa(t),
\end{equation}

The entropy $\mathcal{E}$ serves as an effective metric for assessing the step-wise correctness of \sysname. A conservative threshold is introduced to filter out steps where the agent is likely to perform incorrect actions. Table. \ref{tab:effort_savings} shows the prediction accuracy on the newly collected out-of-domain data can be as high as 76\%. In such cases, a human annotator reviews the surrounding steps near $t$ and provides corrective actions as needed. The agent then resumes execution using the supplied actions until the task is successfully completed.

\subsection{Iterative Annotation and Training}
\label{subsec:iterative}

\begin{figure}[!t]
    \centering
    \begin{subfigure}{0.23\textwidth}  %
        \centering
        \includegraphics[width=\textwidth]{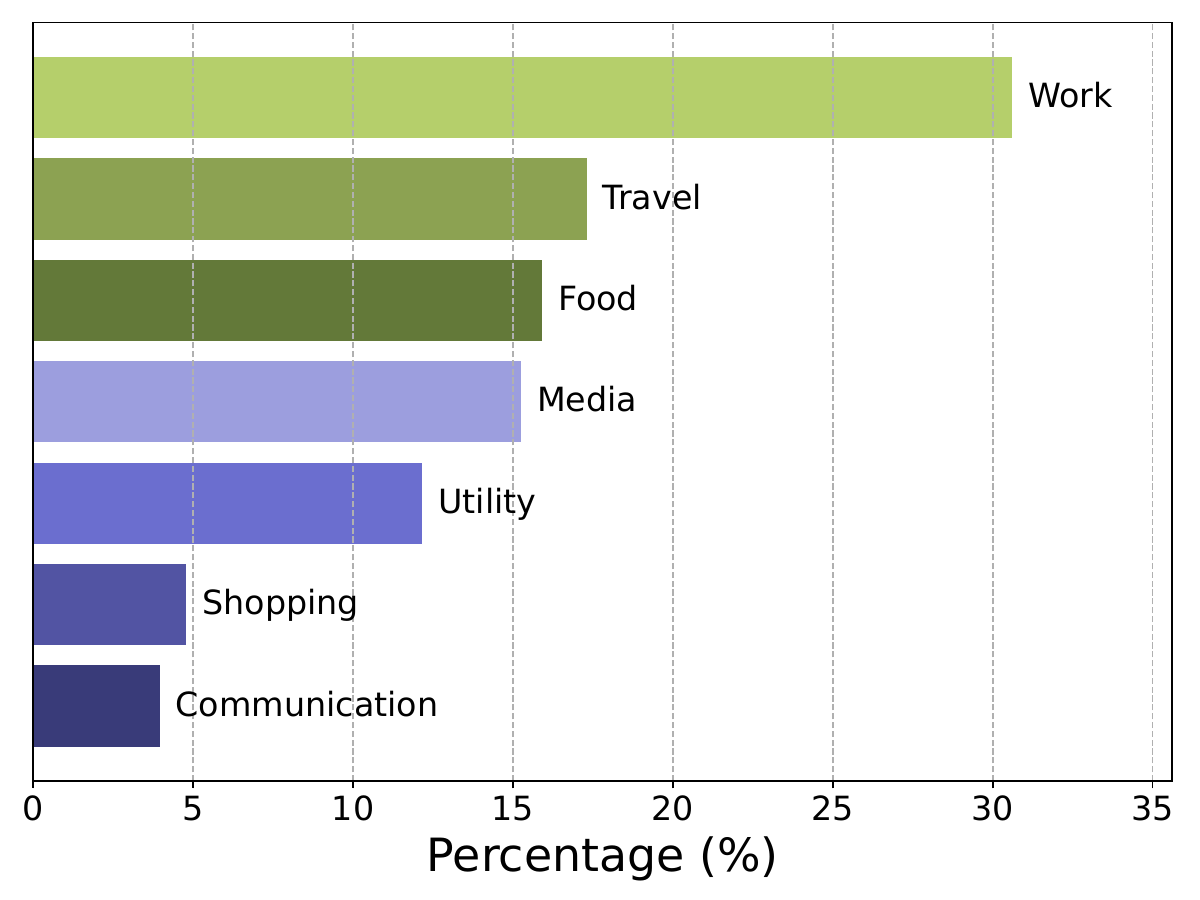}  %
        \caption{\textcolor{black}{Distribution of training data across app. categories}}
        \label{fig:application_categories}
    \end{subfigure}
    \hfill  %
    \begin{subfigure}{0.23\textwidth}  %
        \centering
        \includegraphics[width=\textwidth]{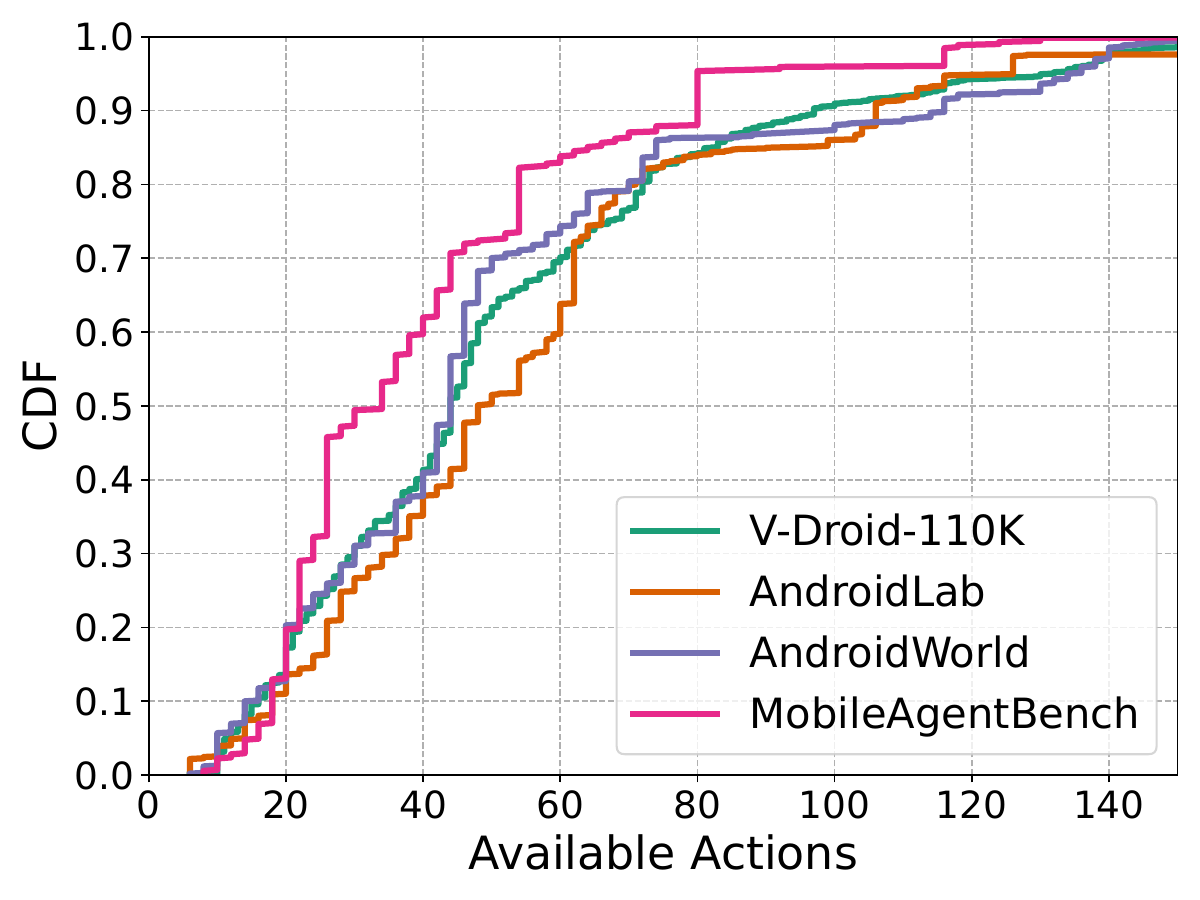}  %
        \caption{\textcolor{black}{Distribution of available actions per step}}
        \label{fig:action_distributions}
    \end{subfigure}
    \caption{\textcolor{black}{Statistics of the collected datasets for training.}}
    \label{fig:dataset_statistics}
\end{figure}

\textcolor{black}{
Leveraging human–agent joint annotation, we iteratively optimize the verifier using $P^3$ training and deploy it to collect new data on out-of-domain tasks and apps over four rounds. To address the cold-start problem in the first round, we manually labeled data without agent assistance. Human annotators selected the correct actions from the extracted action lists provided by \sysname, requiring approximately four minutes per task and about five hours in total to collect the initial 9K data pairs. In subsequent rounds, the agent was deployed to execute tasks and mark stepwise correctness, enabling progressive dataset expansion from 9K to 27K, then to 55K, and finally to 110K after four iterations. Human annotators primarily corrected errors in the trajectories, which reduced the average annotation time to around one minute per task. As shown in Fig.~\ref{fig:dataset_statistics}, the collected dataset spans more than 90 applications across seven common categories.}

\textcolor{black}{As data volume grows, the verifier demonstrates continuous performance improvements in both decision-making capability and annotation accuracy. Table~\ref{tab:effort_savings} presents the truth table across different iterations, comparing verifier predictions based on entropy with step-wise ground truth. The true positive ratio increases from 0.11 to approximately 0.40, while annotation accuracy improves from 0.51 to over 0.70. These improvements indicate that as the verifier’s scoring ability strengthens, the agent correctly annotates more steps, further reducing human annotation effort and enhancing training efficiency.}

\begin{table}[t]
  \caption{\textcolor{black}{Truth table based on entropies of scores. The threshold is set as the median value of the entropies.}}
  \label{tab:effort_savings}
  \begin{tabular}{ccccccc}
    \toprule
    Iterations & Data & TP & TN & FP & FN & Accuracy \\
    \midrule
    Iter. 2 & 27k & 0.11 & 0.40 & 0.39 & 0.10 & 0.51\\
    Iter. 3 & 55k & 0.42 & 0.34 & 0.08 & 0.16 & 0.76\\
    Iter. 4 & 110k & 0.40 & 0.31 & 0.10 & 0.19 & 0.71\\
    \bottomrule
  \end{tabular}
\end{table}

\subsection{Training Details}

In \sysname, the generative verifier is built based on Llama-3.1-8B-Instruct-4-bit with one MLP layer attached to project the token logits into the action scores. We use Q-LoRa training with rank 16. The learning rate is set to 1e-4 with 10 warm-up steps and then gradually reduced to 1e-6 until the end of the training. The training epoch is set to 20 for the 4th iteration, which is larger than the usual LoRa training epoch. The reason is that longer training enlarge the score gap between the correct and rejected actions, which is observed to be more robust during the test time.
The 4th iteration of training is conducted on around 110k data pairs for 90 hours on $16\times$Nvidia A100 40G GPUs.

\begin{figure*}[th]
    \centering
    \begin{subfigure}[b]{0.32\textwidth}  %
        \centering
        \includegraphics[width=\linewidth]{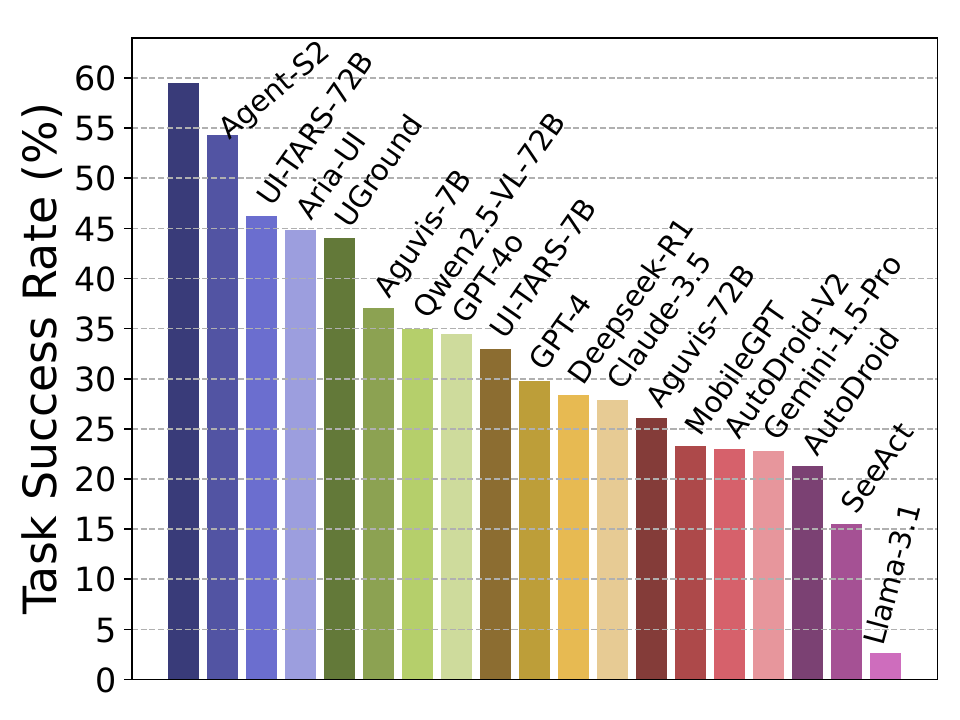}  %
        \caption{AndroidWorld}
        \label{fig:sr_android_world}
    \end{subfigure}
    \hfill  %
    \begin{subfigure}[b]{0.32\textwidth}  %
        \centering
        \includegraphics[width=\linewidth]{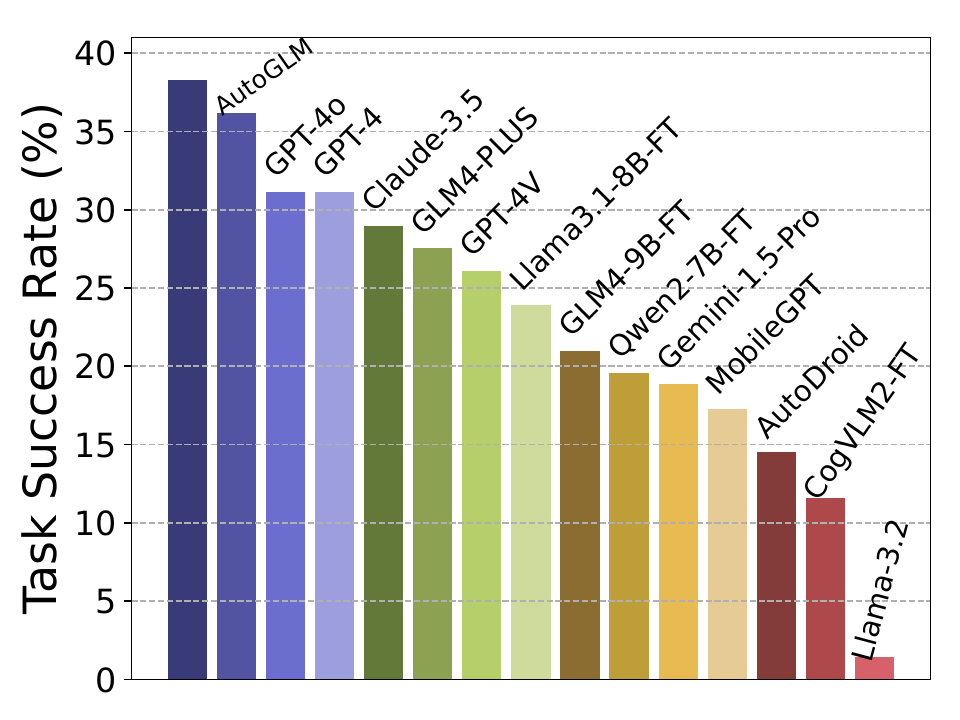}  %
        \caption{AndroidLab}
        \label{fig:sr_android_lab}
    \end{subfigure}
    \hfill  %
    \begin{subfigure}[b]{0.32\textwidth}  %
        \centering
        \includegraphics[width=\linewidth]{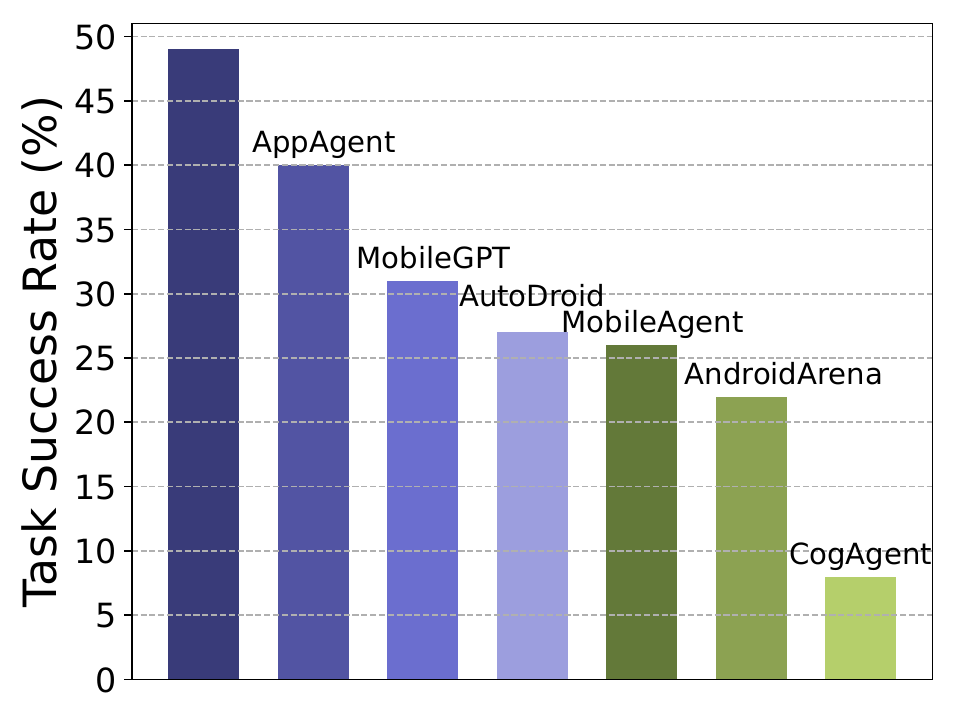}  %
        \caption{MobileAgentBench}
        \label{fig:sr_mobile_agent}
    \end{subfigure}
    \caption{\textcolor{black}{Task success rate achieved by \sysname compared to a range of mobile agents on three public benchmarks. The leftmost bar corresponds to \sysname.}}
    \label{fig:performance comparison}
\end{figure*}

\section{Evaluation}
\label{sec:evaluation}

In the section, we evaluate \sysname to highlight  1) the performance of \sysname compared to SOTA mobile agents in terms of improved task success rate and reduced latency; 2) the training scaling law and the savings of annotation overhead in multiple iterations of annotation and training; and 3) the effectiveness of the training designs and the inference optimizations in \sysname.

\subsection{Experiment Settings}
\subsubsection{Benchmark}
\sysname is evaluated on three widely-used public benchmarks that cross varied types of mobile phones, system versions, applications and user instructions.

\textbf{AndroidWorld}~\cite{rawles2024androidworld} includes 20 APPs, 116 tasks, infinite task instructions that support random initialization to benchmark mobile agents abilities in real-execution environment. It includes challenging tasks that take more than 30 steps to complete and involves multi-app interactions.

\textbf{AndroidLab}~\cite{xu2024androidlab} is a emulation environment that includes 138 tasks in nine kinds of daily used APPs, such as map, calendar, books, music. The agents are required to manage events, edit notes, and check information, etc.

\textcolor{black}{\textbf{MobileAgentBench}~\cite{wang2024mobileagentbench} provides a phone usage environment built on 10 open-source applications and 100 tasks.}

\textcolor{black}{In contrast to static benchmarks, \eg DroidTask~\cite{wen2024autodroid} and AndroidControl~\cite{li2024effects}, the  realistic and dynamic Android environment we evaluate \sysname can better reflect its ability. Among the three benchmarks, AndroidWorld is designated as the in-the-domain test set, whereas AndroidLab and MobileAgentBench are designated as out-of-domain (OOD) test sets. During the training data collection process, we rigorously excluded applications utilized in these two benchmarks to ensure unbiased evaluation. The differences in data distributions between the training set and the test benchmarks are illustrated in Fig.~\ref{fig:action_distributions}.}

\subsubsection{Baselines} \textcolor{black}{\sysname is compared with a wide range of mainstream mobile agents, including text-only agents (T3A \cite{rawles2024androidworld}, AutoDroid \cite{wen2024autodroid}, AutoDroid-V2 \cite{wenAutoDroidV2BoostingSLMbased2024}, MobileGPT \cite{lee2024mobilegpt}, Ponder\&Press \cite{wang2024ponder}), multimodal agents (M3A \cite{rawles2024androidworld}, SeeAct \cite{zheng2024seeact}, AndroidArena \cite{xing2024understanding}, CogAgent \cite{hong2024cogagent}, APPAgent \cite{li2024appagent}, MobileAgent \cite{wang2024mobile}), and agents with grounding models (Agent-S2 \cite{agashe2025agents}, Aria-UI \cite{yang2024aria}, UGround \cite{gou2025uground}, and UI-TARS \cite{qin2025ui}). Advanced LLMs, including GPT-3.5, GPT-4, GPT-4V, GPT-4o, GLM, DeepSeek-R1, Qwen, Llama-3.1, Claude and Gemini, are included as the base models for these agents. All the cloud-source LLMs are prompted using the M3A template in AndroidWorld \cite{rawles2024androidworld}. Note that the baseline differences across benchmarks are attributable to the fact that some agents \cite{liu2024autoglmautonomousfoundationagents, qin2025ui, li2024appagent}, were not open-sourced or lacked reproducible implementations, making it infeasible to evaluate them across all benchmarks. Therefore, we followed the practice and used all baselines available in each benchmark’s leaderboard to ensure fairness and completeness.}

\subsubsection{Metrics} \textcolor{black}{\sysname is evaluated against baselines using two key metrics: task success rate (SR) and latency. We directly adopt the SR number reported in the respective baseline papers on the corresponding benchmarks. To assess latency, we employ two measures: the total step-wise latency of the entire agent and the decision-making latency of the LLM. The total step-wise latency primarily comprises the decision-making latency, the working memory update latency, and the execution time. Latency is evaluated using 20 randomly sampled tasks from benchmarks.}

\subsubsection{Evaluation platform} We evaluate \sysname across various hardware configurations. The agent system runs within an Android emulator on a PC with an Intel i9-10900X CPU. The LLMs used by \sysname and baseline models are tested on NVIDIA GPUs, including 4090, A100, A6000. Unless otherwise specified, all time measurements of \sysname are conducted on a server with two NVIDIA 4090 GPUs.

\begin{figure}[!t]
    \centering
    \begin{subfigure}{0.23\textwidth}  %
        \centering
        \includegraphics[width=\textwidth]{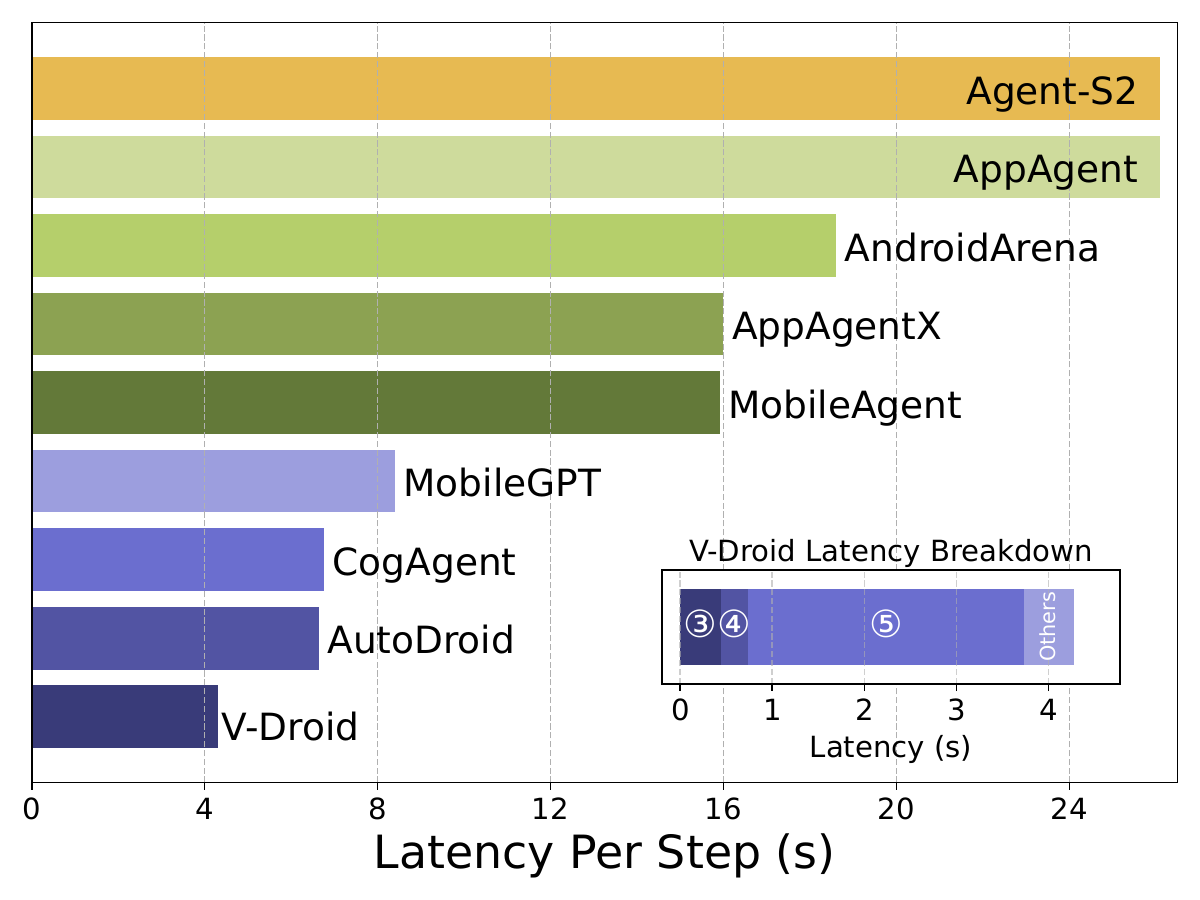}  %
        \caption{Step-wise latency}
        \label{fig:step_latency_android_lab}
    \end{subfigure}
    \hfill  %
    \begin{subfigure}{0.23\textwidth}  %
        \centering
        \includegraphics[width=\textwidth]{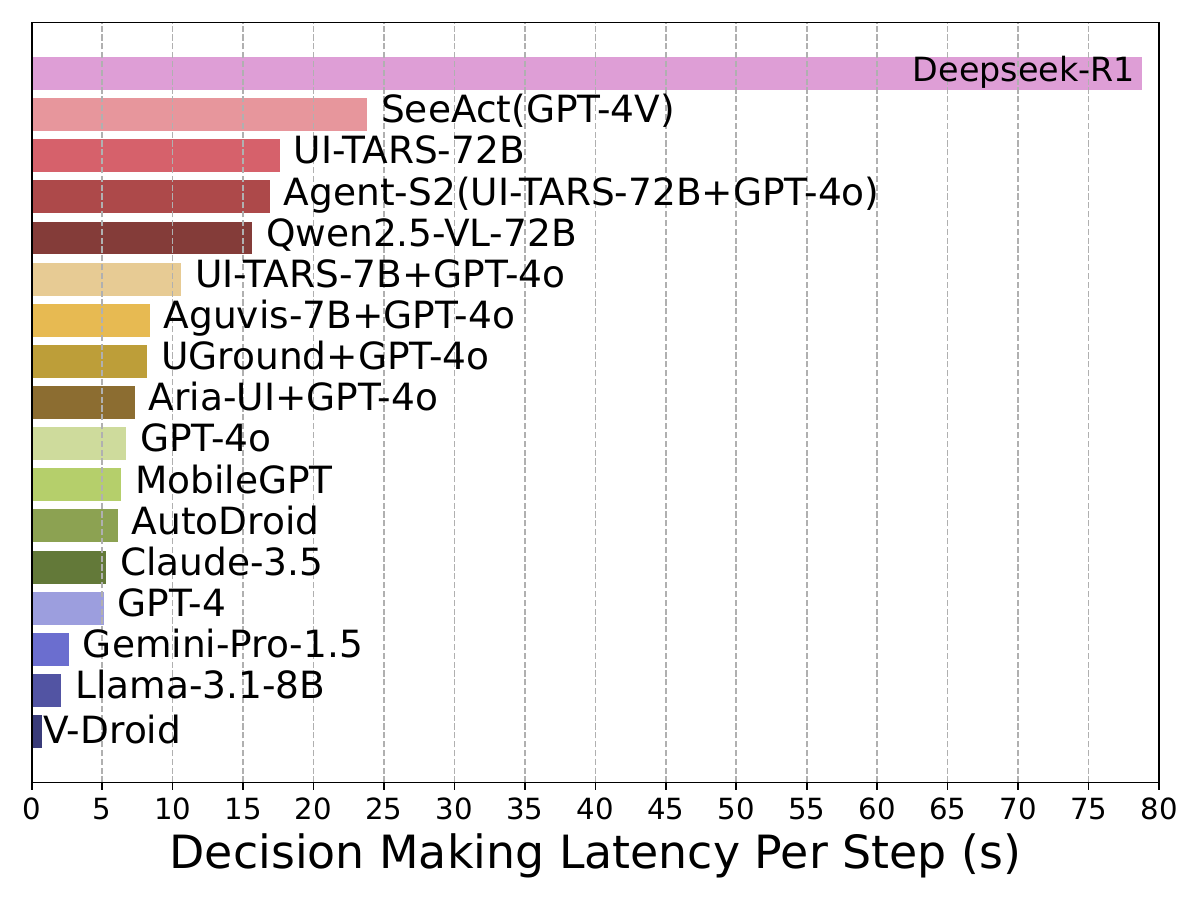}  %
        \caption{Decision-making latency}
        \label{fig:decision_latency_android_world}
    \end{subfigure}
    \caption{\textcolor{black}{The step-wise latency and decision-making latency of \sysname compared to typical mobile agents.}}
    \label{fig:latency}
\end{figure}

\subsection{Performance Improvement}

\textbf{Improved task success rate}. \textcolor{black}{Fig.~\ref{fig:performance comparison} demonstrates that \sysname outperforms existing mobile agents across three realistic MobileAgentBenchmarks, AndroidWorld, AndroidLab, and MobileAgentBench, achieving success rate (SR) improvements of 5.2\%, 2.1\%, and 9.0\%, respectively. Compared to cloud-based LLM-powered agents (\eg GPT-4, GPT-4o, DeepSeek-R1, Gemini-1.5-Pro, Claude-3.5), \sysname achieves 25.0\% and 7.13\% higher SR on AndroidWorld and AndroidLab, respectively. Against advanced mobile agent frameworks (\eg Agent-S2, AutoDroid, AppAgent), \sysname improves SR by 5.2\% on AndroidWorld and 9.0\% on the MobileAgentBench. Compared to models that decompose decision-making into reasoning and grounding (\eg UI-TARS, Aria-UI, UGround, Aguvis), \sysname demonstrates a 14.3\% SR improvement on the AndroidWorld. Against other fine-tuned SLMs (\eg Qwen-VL-7B-FT, Llama-3.1-8B-FT), \sysname achieves a notable 14.3\% improvement on AndroidLab. Compared with memory-driven agents like AutoDroid and MobileGPT, \sysname achieves over 36.2\%, 21.0\%, and 18.0\% SR improvement on the three benchmarks. Note that on MobileAgentBench, baselines such as AppAgent \cite{li2024appagent} already outperform cloud LLMs so those results are not included.}

\textcolor{black}{Unlike existing generation-based GUI agents that operate in continuous UI spaces, \sysname simplifies decision-making by mapping UI states to a finite action space and decomposing the process into verification and completion, making it more tractable for SLMs. In addition, the verifier-driven workflow embeds UI-specific knowledge directly into the action candidates, thereby reducing the complexity of decision-making. Moreover, the $P^3$ training framework, built on large-scale data, further enhances the verifier’s ability to distinguish between similar actions, self-correct errors, and maintain awareness of task progress. }

\textbf{Reduced Latency.} \textcolor{black}{Fig.~\ref{fig:latency} highlights the significant speed advantage of \sysname over SOTA mobile agents in both step-wise and decision-making latency. While SOTA agents typically take over 20 seconds per step, \sysname takes just 4.3 seconds per step, which is $6.1\times$ faster. Specifically, it takes 0.44s for the verification (stage \ding{174} in Fig.~\ref{fig:details of verifier}), 0.30s for the action completion (stage \ding{175}), and 3.03s for the working memory (stage \ding{176}). The action space and the prompt construction takes less than 1ms (stage \ding{172}-\ding{173}) and the photo transition time takes 0.54s on average.}

\textcolor{black}{This efficiency gain stems from \sysname’s verifier-driven workflow for decision-making, which transforms traditional auto-regressive decoding into a parallelized, prefilling-only scoring process. As shown in Fig.~\ref{fig:decision_latency_android_world}, \sysname achieves up to a $32.1\times$ speedup compared to grounding-based agents that decompose decision-making into reasoning and grounding (e.g., UI-TARS, Aria-UI, UGround, Aguvis). Against UI-TARS \cite{qin2025ui} and DeepSeek-R1 with System-2 reasoning \cite{guo2025deepseek}, \sysname achieves $25.1\times$ and $112.2\times$ speed improvements, respectively. To the best of our knowledge, \sysname is the first mobile GUI agent capable of near-real-time decision-making. We further discuss design alternatives for optimizing working memory in Section~\ref{subsec: design alternatives}. }

\textcolor{black}{The action completion in \sysname is only needed in 12.4\% cases. In some input heavy tasks, the stepwise latency of \sysname could be longer. For instance, in calendar-related tasks, there are 25.5\% actions needs completion, which increases the average latency to 0.71s for the completion and 4.7s per step. In contrast, selection-heavy tasks that do not require text input (such as opening Wi-Fi) achieve 4.0s per step.}

\textbf{Comparison with CoT-Disabled Agents.} \textcolor{black}{The advantages of the verification-driven workflow in \sysname are further demonstrated in Table~\ref{tab:cot_comparison} on AndroidWorld. We compare V-Droid with the best agents using cloud-based GPT-4o and open-sourced local-served UI-TARS-7B \cite{qin2025ui} (similar size with \sysname) that support execution with or without CoT. Other agents (e.g., Agent-S2 \cite{agashe2025agents}) rely on mandatory CoT or perform even worse, thus are excluded for this comparison. Disabling CoT reasoning in GPT-4o and UI-TARS leads to reduced decision-making latency. However, this comes at the cost of decreased SR. Despite removing CoT, both GPT-4o and UI-TARS still exhibit higher latency than \sysname, as they autoregressively generate full actions. In contrast, \sysname achieves a substantially higher SR with lower decision-making latency. \sysname requires far fewer input tokens because it operates on simplified XML, lightweight system instructions, and compressed memory, whereas GPT-4o and UI-TARS depend on screenshots and excessive historical context. Besides, GPT-4o’s latency does not scale proportionally with output token reductions, which might due to the internet latency dominate its runtime. These results further underscore the generation–verification gap in GUI agents and highlight the effectiveness of verifier-driven workflow.}

\begin{table}[!t]
\centering
\caption{\textcolor{black}{Decision-making latency of GPT-4o, UI-TARS-7B \cite{qin2025ui} with and without CoT, compared to V-Droid.}}
\label{tab:cot_comparison}
\scalebox{0.85}{
\begin{tabular}{l|cc |cc|c}
\toprule
\textbf{Agent} & \multicolumn{2}{c|}{GPT-4o} & \multicolumn{2}{c|}{UI-TARS-7B} & \sysname \\
\midrule
Decision & w/ CoT & w/o CoT & w/ CoT & w/o CoT & Verifier \\
\midrule
SR (\%) & 41.3 & 39.2 & 33.2  & 29.3 & 59.5 \\
Latency (s) & 6.70 & 5.74 & 10.6 & 0.97 & 0.74 \\
Input Tokens & 6.2K & 6.2K & 8.9K & 2.9K &  2.6K \\
Output Tokens & 63.4  & 19.6 & 75.9 & 14.6 & 54.3 \\
\bottomrule
\end{tabular}
}
\vspace{-0.5em}
\end{table}

\begin{figure}[t]
    \centering
    \includegraphics[width=0.8\linewidth]{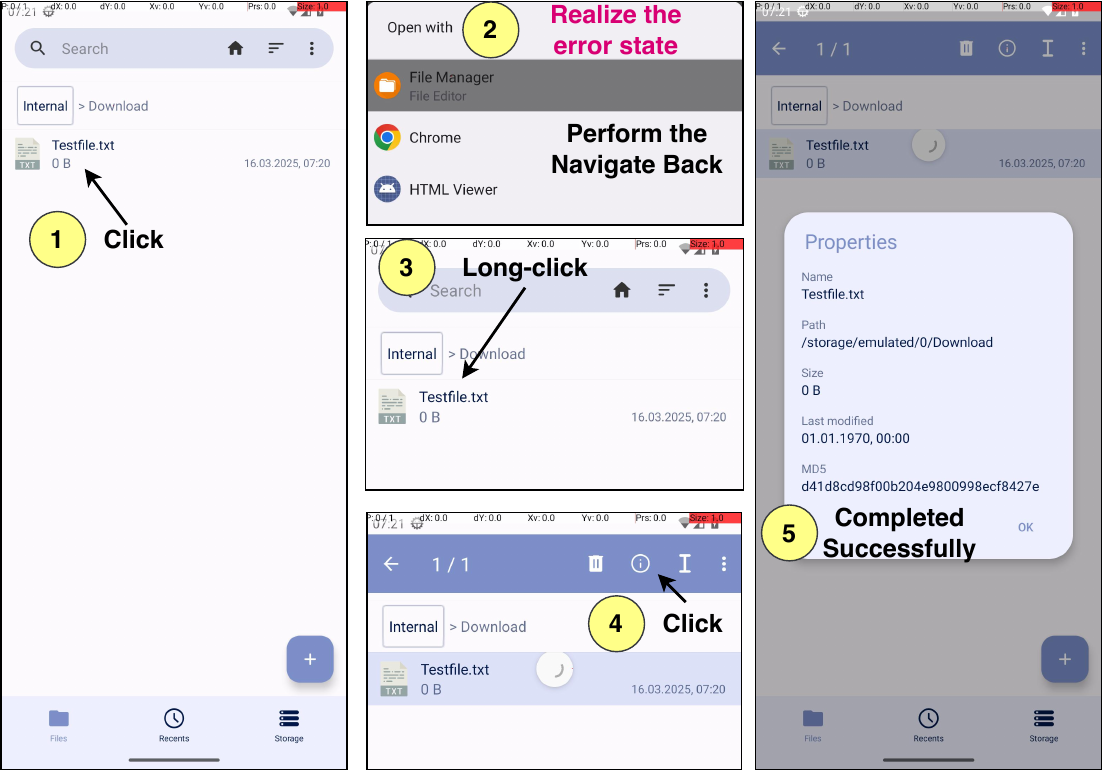}
    \caption{\textcolor{black}{A showcase of V-Droid automating the task "\textit{Check the file property of Textfile.txt}" on MobileAgentBench. Note that both this task and application were excluded from the training process}}
    \label{fig: self-correct demo}
\end{figure}

\textbf{Self-correction showcase.} \textcolor{black}{In Fig. \ref{fig: self-correct demo}, \sysname explores one reasonable action "Click Textfile.txt" but then realizes itself in a wrong status that deviates from the goal. Later, it selects to navigate back and long-press the button to reveal the file properties. Similar cases are observed on the other two benchmarks. Notably, the self-correction training improves the SR of \sysname from 52.2\% to 59.5\% on AndroidWorld and slightly increases the average trajectory length from 10.3 to 11.6 steps due to additional explorations. This modest increase in trajectory length is a reasonable trade-off given the substantial gain in overall success rate.}

\begin{figure}[t]
\centering
    \begin{subfigure}[b]{0.22\textwidth}
    \centering
    \includegraphics[width=\textwidth]{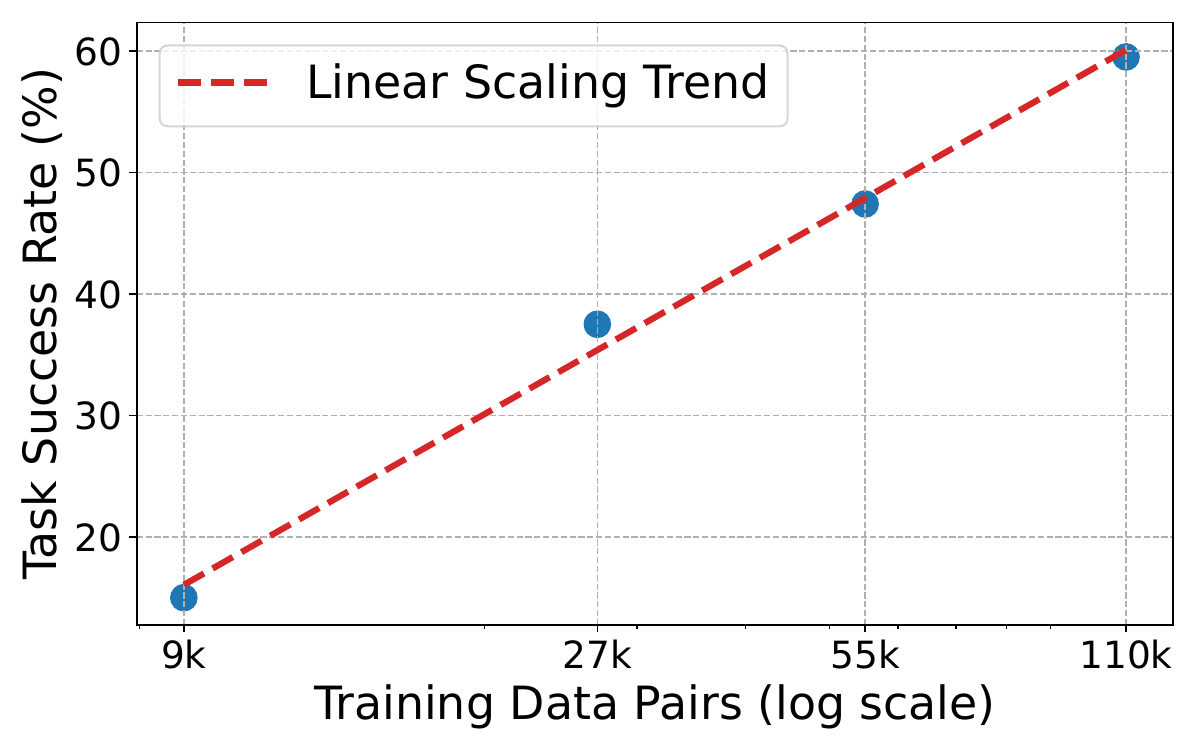}
    \caption{AndroidWorld}
    \label{fig: gain androidworld}
    \end{subfigure}\hfill
    \begin{subfigure}[b]{0.22\textwidth}
    \centering
    \includegraphics[width=\textwidth]{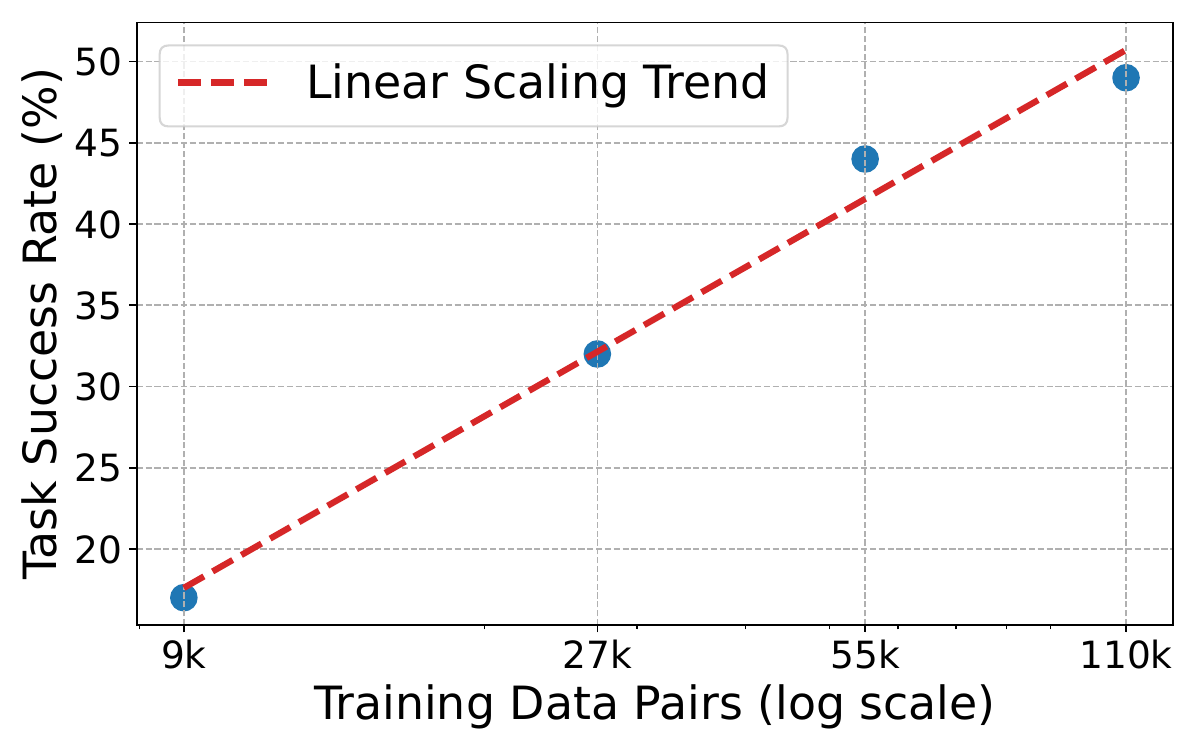}
    \caption{MobileAgentBench}
    \label{fig: gain MobileAgentBench}
    \end{subfigure}
    \caption{Training time scaling law of the generative verifier. As the data pairs scales from 9k to 110k, the performance of the generative verifier boosts.}
    \label{fig: scaling_law}
    \vspace{-1.5em}
\end{figure}

\subsection{Training Scaling Law in \sysname}\label{sec: exp training time scaling}

\textbf{\sysname improves with more training data.} Fig.\ref{fig: scaling_law}(a) and Fig.\ref{fig: scaling_law}(b) illustrate how the \sysname's performance scales with increasing training data. On AndroidWorld, the SR improves from 15.0\% to 37.5\%, then further scales to 47.4\% and 59.5\% as the number of training data pairs increases from 9K (Human-Annotated) to 110K (Human-Agent Joint-Annotated). Similarly, on the MobileAgentBench, the SR increases from 17.0\% to 49.0\% as more training data becomes available. This continuous improvement highlights that \sysname benefits from diverse app environments, instructions, and execution trajectories, which gradually expand throughout the training. The increasing dataset variety enhances the agent’s generalization ability and robustness, leading to more effective decision-making across different tasks.

\begin{figure}[t]
\centering
    \begin{subfigure}[b]{0.23\textwidth}
    \centering
    \includegraphics[width=\textwidth]{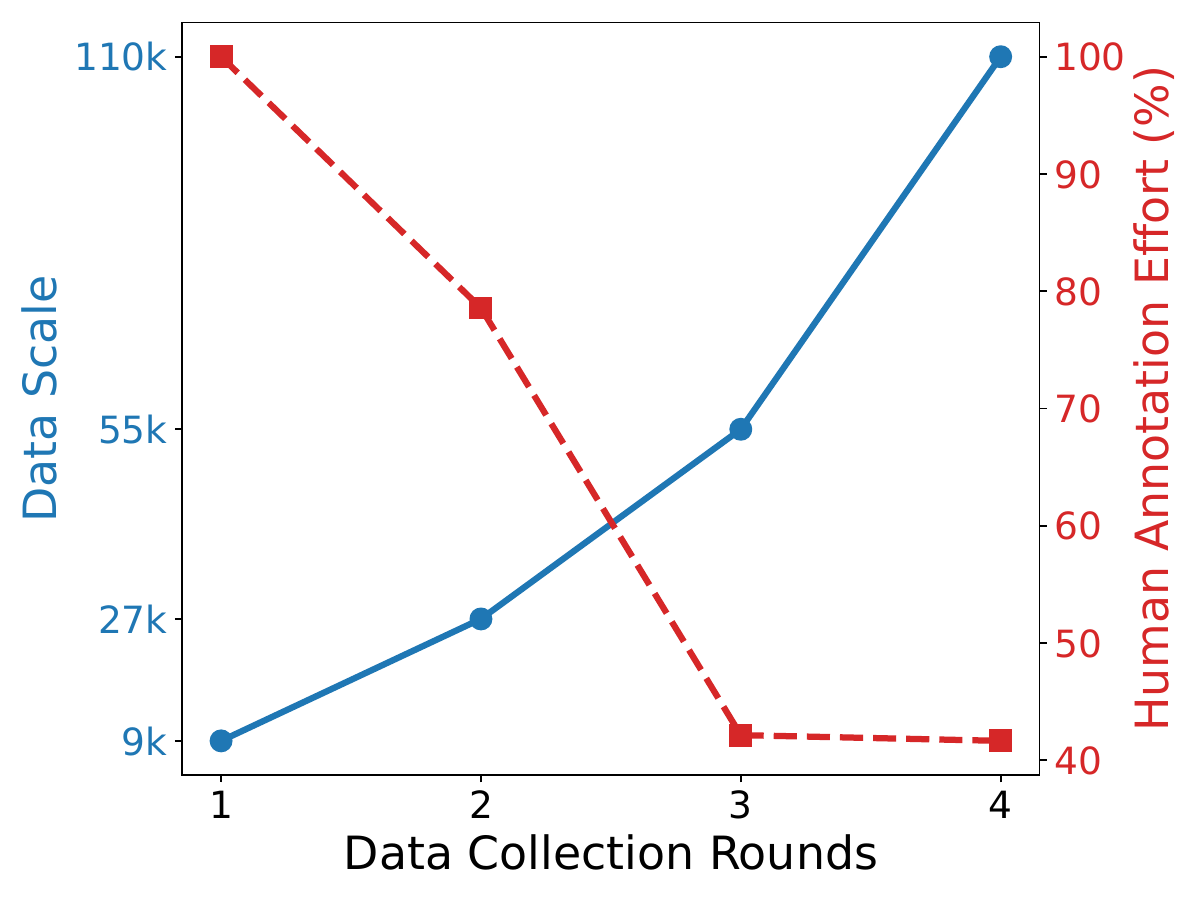}
    \caption{Annotation overhead}
    \label{fig: annotation overhead saved}
    \end{subfigure}
    \begin{subfigure}[b]{0.23\textwidth}
    \centering
    \hfill
    \includegraphics[width=\textwidth]{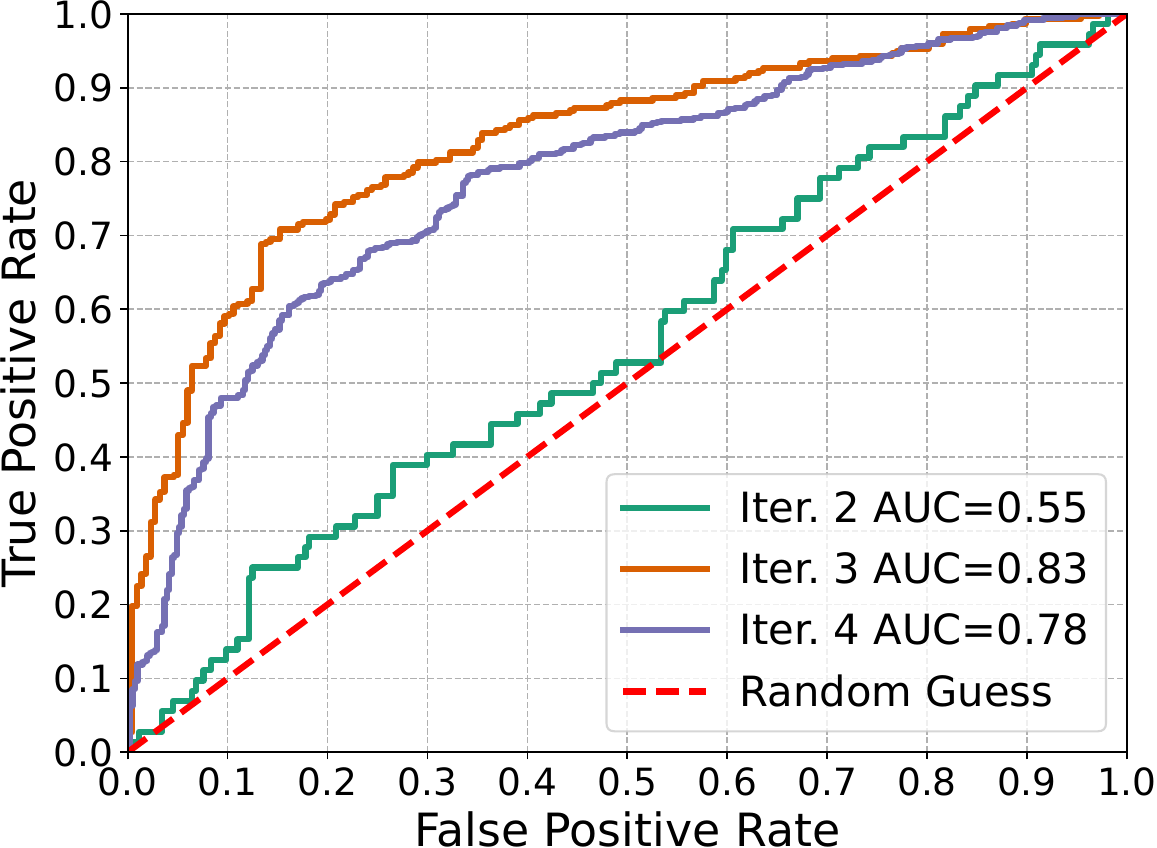}
    \caption{ROC of entropy.}
    \label{fig: roc}
    \end{subfigure}
    \caption{Leveraging Verifier-as-annotator saves the annotation overhead. (a) As the agent ability improves, the annotation overhead of the human annotators decrease. (b) The verifier is an accurate decision classifier after two iterations of training.}
    \label{fig: annotation_overheadthree_subfigures}
\end{figure}

\textbf{Human annotation overhead decreases with better verifier.} As shown in Fig.~\ref{fig: annotation overhead saved}, the human annotation effort, measured as the ratio of data pairs collected by human annotators, gradually decreases across multiple iterative annotation and training cycles. After training with 27K and 55K data pairs, the AUC improves from 0.55 to approximately 0.8, demonstrating a significant enhancement in the verifier’s ability to predict decision correctness. This improved accuracy enables human annotators to quickly rectify errors and recover the data collection process with minimal overhead. To further enhance the verifier’s capabilities, we continue to scale the dataset using the human-agent joint annotation scheme in \sysname, ensuring progressive improvements in annotation efficiency and model performance.

\subsection{Comparison with Design Alternatives} \label{subsec: design alternatives}

\textbf{Architecture and Training Alternatives.} \textcolor{black}{We craft three baselines to justify the designs in \sysname, including \emph{LLM-as-a-Judge}, \emph{Selector} and \emph{Generator}. LLM-as-a-Judge follows the architecture of \sysname, but uses GPT-4 as the verifier. The action score is obtained by extracting the output logits value of "Yes". Selector is presented with XML descriptions and the extracted action lists and is fine-tuned to output the correct action number. Generator follows the architectures of T3A~\cite{rawles2024androidworld} but replaces the policy agent with a fine-tuned LLama-3.1-8B that is trained to output thought chains followed by the chosen actions. We adopt supervised fine-tuning for both designs.}

\begin{table}[t]
    \centering
    \caption{Comparing \sysname with the design alternatives of agent architectures and corresponding training approaches with training data from $1.1$K task steps.}
    \label{tab:design_alternatives}
    \scalebox{1.0}{
    \begin{tabular}{cccc}
        \toprule
        \textbf{Methods} & \textbf{Training} & \textbf{Base Model} & SR (\%) \\
        \midrule
        \sysname & $P^3$ & Llama-3.1-8B & 47.4\\\midrule
        Selector & SFT & Llama-3.1-8B & 35.8\\
        Generator & SFT & Llama-3.1-8B & 27.4 \\
        LLM-as-a-Judge & -- & Llama-3.1-8B & 0\\\midrule
        LLM-as-a-Judge & -- & GPT-4 & 34.5\\
        Generator & -- & GPT-4 & 29.6\\
        \bottomrule
    \end{tabular}
    }
\end{table}

Table~\ref{tab:design_alternatives} highlights the limitations of LLM-As-A-Judge using Llama-3.1-8B without training, which fails to assign accurate action scores, resulting in a 0\% success rate. While GPT-4 performs better due to stronger reasoning and instruction-following abilities, it also outperforms its own generator-based approach, reinforcing the generation-verification gap. However, it is observed that GPT-4 often assigns high scores to multiple actions or low scores to all actions, exposing a misalignment between token space and action space. Training Llama-3.1-8B as either a selector or generator improves success rates to 35.8\% and 27.4\%, respectively. However, these models still underperform compared with \sysname, which leverages a verifier-driven workflow with $P^3$ training. The advantage of $P^3$ training comes from its pair-wise learning structure, where each step with $N$ available actions generates $N-1$ training pairs, effectively amplifying the training data by $N-1$ times. Additionally, pair-wise training enhances the model's ability to distinguish similar UI elements and actions, a crucial factor for accurate action selection.

\textbf{Working memory alternatives}. We observe that the step-wise latency of \sysname is primarily constrained by the time required to construct working memory using an LLM, from 0.7 seconds to 3.8 seconds. It is because we use GPT-4 to update the working memory in current implementation. To mitigate this bottleneck, we explore two alternative designs at test time aimed at reducing latency: 1) \textbf{Action History Only} – This approach retains only a sequential log of past actions as the working memory.
2) \textbf{Rule-based Memory} – This method generates concise, structured descriptions of past actions and UI changes by applying rule-based heuristics. It extracts and compares UI content descriptions before and after an action, enabling a high-level summary. For instance, \textit{Clicked the 'Save' button. Now an 'OK' text box appears, indicating that the action likely succeeded}.

As shown in Table~\ref{tab:working memory alternatives}, using only action history reduces the SR on AndroidWorld to 40.0\%, as the agent struggles to retain key contextual information, leading to repeated actions and suboptimal decisions. Incorporating rule-based memory improves SR to 46.1\%, demonstrating the benefit of structured summaries. However, the SR remains significantly lower than when using LLM-based memory construction, underscoring the importance of high-level action impact summarization and the ability to retain crucial contextual information across steps.

\begin{table}[t]
    \caption{Decision Alternatives on Working Memory.}
    \label{tab:working memory alternatives}
    \centering
    \begin{tabular}{ccc}
        \toprule
        Design Alter. & SR & Latency (s) \\
        \midrule
        LLM-based & 59.5\% & 3.03 \\
        Rule-based  & 46.1\% & 1e-5 \\
        Actions History & 40.0\% & -- \\
        \bottomrule
    \end{tabular}
    \vspace{-1em}
\end{table}

\subsection{System Overhead} \label{sec: inference acceleration}
Running the verifier of \sysname on GPUs requires approximately $16$GB of memory, including the KV cache. Across the evaluated benchmarks, \sysname verifies an average of $50.3$ actions per task step.

Such verifications could be significantly accelerated with the batching inference and prefix caching. We highlight the decision-making latency on $4\times$ NVIDIA GTX A100 80G, $4\times$ NVIDIA GTX A6000, and $2\times$ NVIDIA GTX 4090.

As shown in Table \ref{tab:gpu_performance}, given the optimization of the prompt format in \sysname that maximizes the length of the shared prefix, a $10\times$ speed up is obtained from the prefix caching across different actions, steps, and tasks. Furthermore, \sysname turns the auto-regressive decision making scheme of LLM agents into the parallel pre-filling only verification scheme, which can be conducted in batch on multiple GPUs. The parallelism further decreases the selection time to 0.42s, 0.44s, and 0.52s with $4\times$ A100, $2\times$ 4090, and $4\times$ A6000.

\textcolor{black}{We also try to infer the overhead running the verifier on mobile devices. Taking the shared prefix into account, the total input token count for all action verifications per step is approximately 1.1K tokens. Given a prefill speed of 450 tokens per second on the Qualcomm Snapdragon 8 Gen 3 NPU~\cite{powerserve2024}, the decision-making latency is around 2.5 seconds.}

\begin{table}[t]
    \centering
    \caption{Decision-making latency with and without the prefix caching on different instances of 4090, A100 and A6000 GPUs.} 
    \label{tab:gpu_performance}
    \begin{adjustbox}{width=\linewidth}
    \begin{tabular}{lcccccccccccc}
        \toprule
        \textbf{GPU} & \multicolumn{4}{c}{\textbf{A100}} & \multicolumn{3}{c}{\textbf{4090}} & \multicolumn{4}{c}{\textbf{A6000}} \\
        \cmidrule(lr){2-5} \cmidrule(lr){6-8} \cmidrule(lr){9-12}
        \textbf{Num.} & 4 & 2 & 1 & 1 & 2 & 1 & 1 & 4 & 2 & 1 & 1 \\
        \textbf{P.C.}  & W/ & W/ & W/ & W/O & W/ & W/ & W/O & W/ & W/ & W/ & W/O \\
        \textbf{Latecy (s)} & 0.717 & 0.932 & 1.446 & 7.116 & 0.744 & 1.034 & 7.847 & 0.808 & 1.027 & 1.555 & 11.36 \\
        \bottomrule
    \end{tabular}

    \end{adjustbox}  
\end{table}

    \subsection{\textcolor{black}{Failure Study}}

    \textcolor{black}{We conducted a manual analysis of the failure cases for \sysname and MobileGPT across more than 300 tasks spanning three benchmarks. We identify four categories of failure: 1) Hallucinated decisions, where erroneous actions are taken despite the agent possessing accurate memory and full observational input; 2) Inaccurate memory, where erroneous contextual memory leads the agent to make wrong decisions; 3) Incomplete actions, wherein available action types are insufficient to complete the task; and 4) Modality limitations, where tasks that demands vision capabilities cannot be accomplished by text-only LLM agents. MobileAgentBench tasks inherently require minimal perception and memory, leading to a high percentage of failures in the hallucinated decision category.}

\begin{figure}[t]
    \centering
    \includegraphics[width=0.6\linewidth]{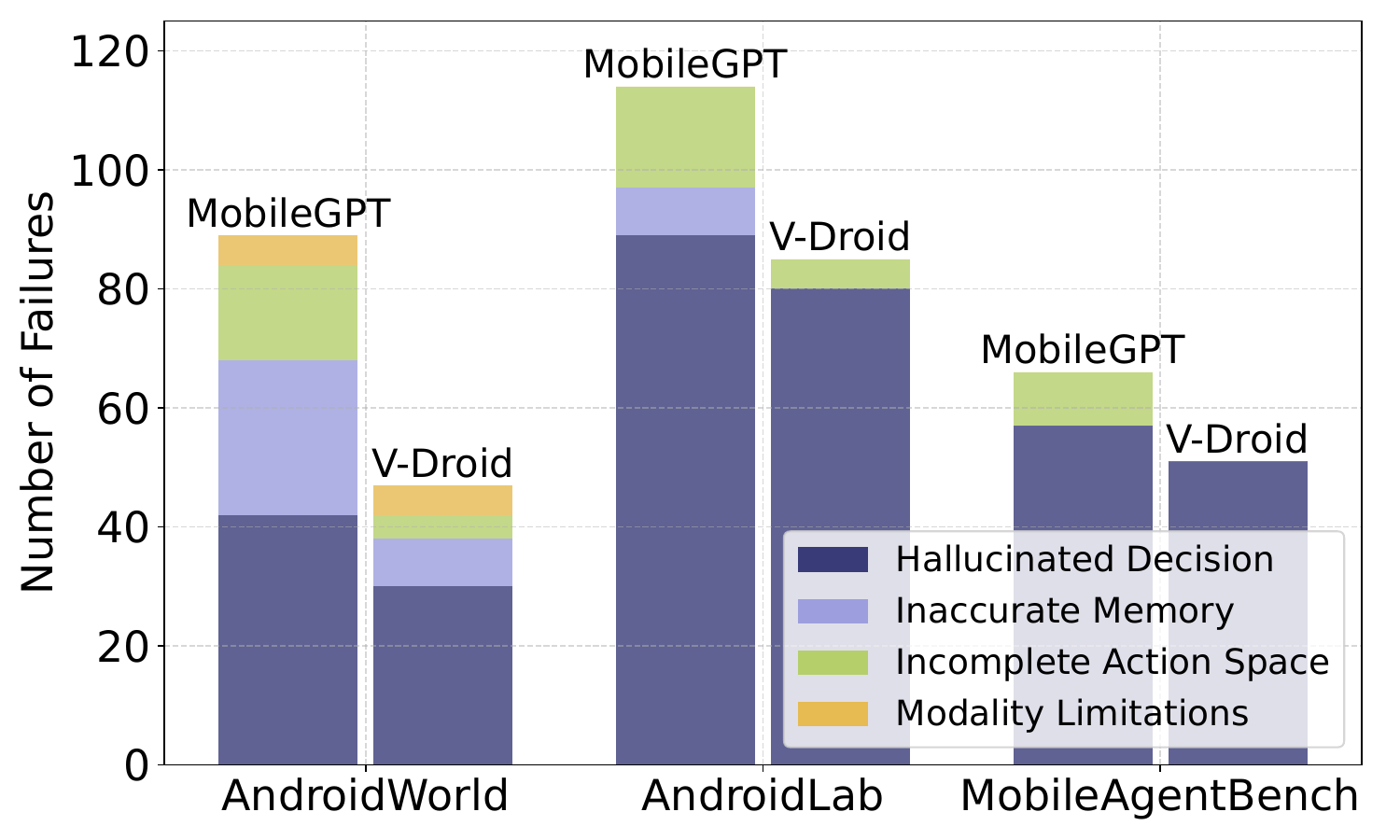}
    \caption{\textcolor{black}{Failure Analysis of MobileGPT and \sysname.}}
    \label{fig:failure_analysis}
\end{figure}

    \textcolor{black}{
    As illustrated in Fig.~\ref{fig:failure_analysis}, failure cases of MobileGPT primarily stem from suboptimal decision-making. For example, given the task instruction \textit{Add one expense (\$307.01, in Health Care categories) to APP Expense Pro}, MobileGPT erroneously inputs the wrong amount and neglects to select the \textit{Health Care} category, whereas \sysname accurately provides all required information. There are still a large portion of failure cases of \sysname in hallucinated decisions. For example, when required to share a file, \sysname presses the file to read it instead of long-pressing it to reveal the \textit{sharing} button, which might due to the lack of functional understanding of some UI elements. This observation further underscores the necessity of a larger scale training on more diverse tasks. Additional sources of failure of \sysname include an incomplete action space caused by inaccurate information from the accessibility service and the lack of visual processing required for interpreting images and videos, which are further discussed in \S~\ref{sec:discussion}.}

\section{Discussion}
\label{sec:discussion}

\textbf{Multimodality}. \textcolor{black}{
Although \sysname is currently text-only, the proposed approach can also be extended to train multimodal mobile agents. For instance, \sysname can be integrated with a grounding model with vision capability, which generate the initial actions for \sysname instead of relying on accessibility services, followed by the action verification and completion process. We also plan to further train a vision-language model as the verifier as the future work, which assigns scores to actions based on the vision and text information jointly.}

\textbf{Security and privacy}. In \sysname, we adopt methods similar to those proposed in~\cite{wen2024autodroid} to conduct security and privacy checks for actions prior to execution. Additionally, $P^3$ training offers a unique opportunity to train \sysname to adhere to security guidelines, which will be explored as part of our future work.

\section{Conclusion}

\textcolor{black}{
For the first time, \sysname demonstrates near-real-time, effective decision-making for mobile agents. It discretizes the decision space for mobile interactions into a finite set of action candidates, transforming the autoregressive process of generator-based agents into a parallelized, prefilling-only scoring mechanism using a generative verifier. $P^3$ training and the scalable human-agent joint annotation framework, significantly enhances the verifier’s decision-making capabilities. Experimental results showcase the training scaling law of the verifier-driven architecture for mobile task automation.
}

\begin{acks}
This research is partially supported by Singapore Ministry of Education under its AcRF Tier 1 grant RT14/22, the Global STEM Professorship Scheme of Hong Kong, the HKUST start up grant, and the Research Grants Council (RGC) General Research Fund (GRF) 16210425.
\end{acks}

\balance
\bibliographystyle{ACM-Reference-Format}
\bibliography{ref}


\begin{thebibliography}{39}


\ifx \showCODEN    \undefined \def \showCODEN     #1{\unskip}     \fi
\ifx \showISBNx    \undefined \def \showISBNx     #1{\unskip}     \fi
\ifx \showISBNxiii \undefined \def \showISBNxiii  #1{\unskip}     \fi
\ifx \showISSN     \undefined \def \showISSN      #1{\unskip}     \fi
\ifx \showLCCN     \undefined \def \showLCCN      #1{\unskip}     \fi
\ifx \shownote     \undefined \def \shownote      #1{#1}          \fi
\ifx \showarticletitle \undefined \def \showarticletitle #1{#1}   \fi
\ifx \showURL      \undefined \def \showURL       {\relax}        \fi
\providecommand\bibfield[2]{#2}
\providecommand\bibinfo[2]{#2}
\providecommand\natexlab[1]{#1}
\providecommand\showeprint[2][]{arXiv:#2}

\bibitem[Agashe et~al\mbox{.}(2025)]%
        {agashe2025agents}
\bibfield{author}{\bibinfo{person}{Saaket Agashe}, \bibinfo{person}{Jiuzhou Han}, \bibinfo{person}{Shuyu Gan}, \bibinfo{person}{Jiachen Yang}, \bibinfo{person}{Ang Li}, {and} \bibinfo{person}{Xin~Eric Wang}.} \bibinfo{year}{2025}\natexlab{}.
\newblock \showarticletitle{{Agent S: An Open Agentic Framework that Uses Computers Like a Human}}. In \bibinfo{booktitle}{\emph{International Conference on Learning Representations (ICLR)}}.
\newblock
\urldef\tempurl%
\url{https://arxiv.org/abs/2410.08164}
\showURL{%
\tempurl}


\bibitem[{Android Developers}(2025a)]%
        {Android_AccessibilityWindowInfo}
\bibfield{author}{\bibinfo{person}{{Android Developers}}.} \bibinfo{year}{2025}\natexlab{a}.
\newblock \bibinfo{title}{AccessibilityWindowInfo}.
\newblock \bibinfo{howpublished}{\url{https://developer.android.com/reference/android/view/accessibility/AccessibilityWindowInfo}}.
\newblock
\newblock
\shownote{Accessed: 2025-03-05}.


\bibitem[{Android Developers}(2025b)]%
        {Android_Command}
\bibfield{author}{\bibinfo{person}{{Android Developers}}.} \bibinfo{year}{2025}\natexlab{b}.
\newblock \bibinfo{title}{Android Debug Bridge}.
\newblock \bibinfo{howpublished}{\url{https://developer.android.com/tools/adb}}.
\newblock
\newblock
\shownote{Accessed: 2025-03-05}.


\bibitem[Chen and Li(2024)]%
        {chen2024octopus}
\bibfield{author}{\bibinfo{person}{Wei Chen} {and} \bibinfo{person}{Zhiyuan Li}.} \bibinfo{year}{2024}\natexlab{}.
\newblock \showarticletitle{Octopus v2: On-device language model for super agent}.
\newblock \bibinfo{journal}{\emph{arXiv preprint arXiv:2404.01744}} (\bibinfo{year}{2024}).
\newblock


\bibitem[Ding(2024)]%
        {dingMobileAgentEnhancingMobile2024}
\bibfield{author}{\bibinfo{person}{Tinghe Ding}.} \bibinfo{year}{2024}\natexlab{}.
\newblock \bibinfo{title}{{{MobileAgent}}: Enhancing Mobile Control via Human-Machine Interaction and {{SOP}} Integration}.
\newblock
\showeprint[arxiv]{2401.04124}~[cs]


\bibitem[Gao et~al\mbox{.}(2024)]%
        {gao2024mobileviewslargescalemobilegui}
\bibfield{author}{\bibinfo{person}{Longxi Gao}, \bibinfo{person}{Li Zhang}, \bibinfo{person}{Shihe Wang}, \bibinfo{person}{Shangguang Wang}, \bibinfo{person}{Yuanchun Li}, {and} \bibinfo{person}{Mengwei Xu}.} \bibinfo{year}{2024}\natexlab{}.
\newblock \bibinfo{title}{MobileViews: A Large-Scale Mobile GUI Dataset}.
\newblock
\showeprint[arxiv]{2409.14337}~[cs.HC]
\urldef\tempurl%
\url{https://arxiv.org/abs/2409.14337}
\showURL{%
\tempurl}


\bibitem[Gou et~al\mbox{.}(2025)]%
        {gou2025uground}
\bibfield{author}{\bibinfo{person}{Boyu Gou}, \bibinfo{person}{Ruohan Wang}, \bibinfo{person}{Boyuan Zheng}, \bibinfo{person}{Yanan Xie}, \bibinfo{person}{Cheng Chang}, \bibinfo{person}{Yiheng Shu}, \bibinfo{person}{Huan Sun}, {and} \bibinfo{person}{Yu Su}.} \bibinfo{year}{2025}\natexlab{}.
\newblock \showarticletitle{Navigating the Digital World as Humans Do: Universal Visual Grounding for {GUI} Agents}. In \bibinfo{booktitle}{\emph{The Thirteenth International Conference on Learning Representations}}.
\newblock
\urldef\tempurl%
\url{https://openreview.net/forum?id=kxnoqaisCT}
\showURL{%
\tempurl}


\bibitem[Guo et~al\mbox{.}(2025)]%
        {guo2025deepseek}
\bibfield{author}{\bibinfo{person}{Daya Guo}, \bibinfo{person}{Dejian Yang}, \bibinfo{person}{Haowei Zhang}, \bibinfo{person}{Junxiao Song}, \bibinfo{person}{Ruoyu Zhang}, \bibinfo{person}{Runxin Xu}, \bibinfo{person}{Qihao Zhu}, \bibinfo{person}{Shirong Ma}, \bibinfo{person}{Peiyi Wang}, \bibinfo{person}{Xiao Bi}, {et~al\mbox{.}}} \bibinfo{year}{2025}\natexlab{}.
\newblock \showarticletitle{Deepseek-r1: Incentivizing reasoning capability in llms via reinforcement learning}.
\newblock \bibinfo{journal}{\emph{arXiv preprint arXiv:2501.12948}} (\bibinfo{year}{2025}).
\newblock


\bibitem[Hong et~al\mbox{.}(2024)]%
        {hong2024cogagent}
\bibfield{author}{\bibinfo{person}{Wenyi Hong}, \bibinfo{person}{Weihan Wang}, \bibinfo{person}{Qingsong Lv}, \bibinfo{person}{Jiazheng Xu}, \bibinfo{person}{Wenmeng Yu}, \bibinfo{person}{Junhui Ji}, \bibinfo{person}{Yan Wang}, \bibinfo{person}{Zihan Wang}, \bibinfo{person}{Yuxiao Dong}, \bibinfo{person}{Ming Ding}, {et~al\mbox{.}}} \bibinfo{year}{2024}\natexlab{}.
\newblock \showarticletitle{Cogagent: A visual language model for gui agents}. In \bibinfo{booktitle}{\emph{Proceedings of the IEEE/CVF Conference on Computer Vision and Pattern Recognition}}. \bibinfo{pages}{14281--14290}.
\newblock


\bibitem[Kwon et~al\mbox{.}(2023)]%
        {kwon2023efficientmemorymanagementlarge}
\bibfield{author}{\bibinfo{person}{Woosuk Kwon}, \bibinfo{person}{Zhuohan Li}, \bibinfo{person}{Siyuan Zhuang}, \bibinfo{person}{Ying Sheng}, \bibinfo{person}{Lianmin Zheng}, \bibinfo{person}{Cody~Hao Yu}, \bibinfo{person}{Joseph~E. Gonzalez}, \bibinfo{person}{Hao Zhang}, {and} \bibinfo{person}{Ion Stoica}.} \bibinfo{year}{2023}\natexlab{}.
\newblock \bibinfo{title}{Efficient Memory Management for Large Language Model Serving with PagedAttention}.
\newblock
\showeprint[arxiv]{2309.06180}~[cs.LG]
\urldef\tempurl%
\url{https://arxiv.org/abs/2309.06180}
\showURL{%
\tempurl}


\bibitem[Lee et~al\mbox{.}(2024)]%
        {lee2024mobilegpt}
\bibfield{author}{\bibinfo{person}{Sunjae Lee}, \bibinfo{person}{Junyoung Choi}, \bibinfo{person}{Jungjae Lee}, \bibinfo{person}{Munim~Hasan Wasi}, \bibinfo{person}{Hojun Choi}, \bibinfo{person}{Steve Ko}, \bibinfo{person}{Sangeun Oh}, {and} \bibinfo{person}{Insik Shin}.} \bibinfo{year}{2024}\natexlab{}.
\newblock \showarticletitle{Mobilegpt: Augmenting llm with human-like app memory for mobile task automation}. In \bibinfo{booktitle}{\emph{Proceedings of the 30th Annual International Conference on Mobile Computing and Networking}}. \bibinfo{pages}{1119--1133}.
\newblock


\bibitem[Li et~al\mbox{.}(2023)]%
        {li2023api}
\bibfield{author}{\bibinfo{person}{Minghao Li}, \bibinfo{person}{Yingxiu Zhao}, \bibinfo{person}{Bowen Yu}, \bibinfo{person}{Feifan Song}, \bibinfo{person}{Hangyu Li}, \bibinfo{person}{Haiyang Yu}, \bibinfo{person}{Zhoujun Li}, \bibinfo{person}{Fei Huang}, {and} \bibinfo{person}{Yongbin Li}.} \bibinfo{year}{2023}\natexlab{}.
\newblock \showarticletitle{Api-bank: A comprehensive benchmark for tool-augmented llms}.
\newblock \bibinfo{journal}{\emph{arXiv preprint arXiv:2304.08244}} (\bibinfo{year}{2023}).
\newblock


\bibitem[Li et~al\mbox{.}(2024a)]%
        {li2024effects}
\bibfield{author}{\bibinfo{person}{Wei Li}, \bibinfo{person}{William Bishop}, \bibinfo{person}{Alice Li}, \bibinfo{person}{Chris Rawles}, \bibinfo{person}{Folawiyo Campbell-Ajala}, \bibinfo{person}{Divya Tyamagundlu}, {and} \bibinfo{person}{Oriana Riva}.} \bibinfo{year}{2024}\natexlab{a}.
\newblock \showarticletitle{On the Effects of Data Scale on Computer Control Agents}.
\newblock \bibinfo{journal}{\emph{arXiv preprint arXiv:2406.03679}} (\bibinfo{year}{2024}).
\newblock


\bibitem[Li et~al\mbox{.}(2024b)]%
        {li2024appagent}
\bibfield{author}{\bibinfo{person}{Yanda Li}, \bibinfo{person}{Chi Zhang}, \bibinfo{person}{Wanqi Yang}, \bibinfo{person}{Bin Fu}, \bibinfo{person}{Pei Cheng}, \bibinfo{person}{Xin Chen}, \bibinfo{person}{Ling Chen}, {and} \bibinfo{person}{Yunchao Wei}.} \bibinfo{year}{2024}\natexlab{b}.
\newblock \showarticletitle{Appagent v2: Advanced agent for flexible mobile interactions}.
\newblock \bibinfo{journal}{\emph{arXiv preprint arXiv:2408.11824}} (\bibinfo{year}{2024}).
\newblock


\bibitem[Lightman et~al\mbox{.}(2023)]%
        {lightmanLetsVerifyStep2023}
\bibfield{author}{\bibinfo{person}{Hunter Lightman}, \bibinfo{person}{Vineet Kosaraju}, \bibinfo{person}{Yura Burda}, \bibinfo{person}{Harri Edwards}, \bibinfo{person}{Bowen Baker}, \bibinfo{person}{Teddy Lee}, \bibinfo{person}{Jan Leike}, \bibinfo{person}{John Schulman}, \bibinfo{person}{Ilya Sutskever}, {and} \bibinfo{person}{Karl Cobbe}.} \bibinfo{year}{2023}\natexlab{}.
\newblock \bibinfo{title}{Let's {{Verify Step}} by {{Step}}}.
\newblock
\showeprint[arxiv]{2305.20050}


\bibitem[Liu et~al\mbox{.}(2024)]%
        {liu2024autoglmautonomousfoundationagents}
\bibfield{author}{\bibinfo{person}{Xiao Liu}, \bibinfo{person}{Bo Qin}, \bibinfo{person}{Dongzhu Liang}, \bibinfo{person}{Guang Dong}, \bibinfo{person}{Hanyu Lai}, \bibinfo{person}{Hanchen Zhang}, \bibinfo{person}{Hanlin Zhao}, \bibinfo{person}{Iat~Long Iong}, \bibinfo{person}{Jiadai Sun}, \bibinfo{person}{Jiaqi Wang}, \bibinfo{person}{Junjie Gao}, \bibinfo{person}{Junjun Shan}, \bibinfo{person}{Kangning Liu}, \bibinfo{person}{Shudan Zhang}, \bibinfo{person}{Shuntian Yao}, \bibinfo{person}{Siyi Cheng}, \bibinfo{person}{Wentao Yao}, \bibinfo{person}{Wenyi Zhao}, \bibinfo{person}{Xinghan Liu}, \bibinfo{person}{Xinyi Liu}, \bibinfo{person}{Xinying Chen}, \bibinfo{person}{Xinyue Yang}, \bibinfo{person}{Yang Yang}, \bibinfo{person}{Yifan Xu}, \bibinfo{person}{Yu Yang}, \bibinfo{person}{Yujia Wang}, \bibinfo{person}{Yulin Xu}, \bibinfo{person}{Zehan Qi}, \bibinfo{person}{Yuxiao Dong}, {and} \bibinfo{person}{Jie Tang}.} \bibinfo{year}{2024}\natexlab{}.
\newblock \bibinfo{title}{AutoGLM: Autonomous Foundation Agents for GUIs}.
\newblock
\showeprint[arxiv]{2411.00820}~[cs.HC]
\urldef\tempurl%
\url{https://arxiv.org/abs/2411.00820}
\showURL{%
\tempurl}


\bibitem[Project(2024)]%
        {powerserve2024}
\bibfield{author}{\bibinfo{person}{PowerServe Project}.} \bibinfo{year}{2024}\natexlab{}.
\newblock \bibinfo{title}{PowerServe: Efficient LLM Serving}.
\newblock \bibinfo{howpublished}{\url{https://github.com/powerserve-project/PowerServe}}.
\newblock
\newblock
\shownote{Accessed: March 19, 2025}.


\bibitem[Qin et~al\mbox{.}(2025)]%
        {qin2025ui}
\bibfield{author}{\bibinfo{person}{Yujia Qin}, \bibinfo{person}{Yining Ye}, \bibinfo{person}{Junjie Fang}, \bibinfo{person}{Haoming Wang}, \bibinfo{person}{Shihao Liang}, \bibinfo{person}{Shizuo Tian}, \bibinfo{person}{Junda Zhang}, \bibinfo{person}{Jiahao Li}, \bibinfo{person}{Yunxin Li}, \bibinfo{person}{Shijue Huang}, {et~al\mbox{.}}} \bibinfo{year}{2025}\natexlab{}.
\newblock \showarticletitle{UI-TARS: Pioneering Automated GUI Interaction with Native Agents}.
\newblock \bibinfo{journal}{\emph{arXiv preprint arXiv:2501.12326}} (\bibinfo{year}{2025}).
\newblock


\bibitem[Qwen et~al\mbox{.}(2025)]%
        {qwen2025qwen25technicalreport}
\bibfield{author}{\bibinfo{person}{Qwen}, \bibinfo{person}{:}, \bibinfo{person}{An Yang}, \bibinfo{person}{Baosong Yang}, \bibinfo{person}{Beichen Zhang}, \bibinfo{person}{Binyuan Hui}, \bibinfo{person}{Bo Zheng}, \bibinfo{person}{Bowen Yu}, \bibinfo{person}{Chengyuan Li}, \bibinfo{person}{Dayiheng Liu}, \bibinfo{person}{Fei Huang}, \bibinfo{person}{Haoran Wei}, \bibinfo{person}{Huan Lin}, \bibinfo{person}{Jian Yang}, \bibinfo{person}{Jianhong Tu}, \bibinfo{person}{Jianwei Zhang}, \bibinfo{person}{Jianxin Yang}, \bibinfo{person}{Jiaxi Yang}, \bibinfo{person}{Jingren Zhou}, \bibinfo{person}{Junyang Lin}, \bibinfo{person}{Kai Dang}, \bibinfo{person}{Keming Lu}, \bibinfo{person}{Keqin Bao}, \bibinfo{person}{Kexin Yang}, \bibinfo{person}{Le Yu}, \bibinfo{person}{Mei Li}, \bibinfo{person}{Mingfeng Xue}, \bibinfo{person}{Pei Zhang}, \bibinfo{person}{Qin Zhu}, \bibinfo{person}{Rui Men}, \bibinfo{person}{Runji Lin}, \bibinfo{person}{Tianhao Li}, \bibinfo{person}{Tianyi Tang}, \bibinfo{person}{Tingyu Xia},
  \bibinfo{person}{Xingzhang Ren}, \bibinfo{person}{Xuancheng Ren}, \bibinfo{person}{Yang Fan}, \bibinfo{person}{Yang Su}, \bibinfo{person}{Yichang Zhang}, \bibinfo{person}{Yu Wan}, \bibinfo{person}{Yuqiong Liu}, \bibinfo{person}{Zeyu Cui}, \bibinfo{person}{Zhenru Zhang}, {and} \bibinfo{person}{Zihan Qiu}.} \bibinfo{year}{2025}\natexlab{}.
\newblock \bibinfo{title}{Qwen2.5 Technical Report}.
\newblock
\showeprint[arxiv]{2412.15115}~[cs.CL]
\urldef\tempurl%
\url{https://arxiv.org/abs/2412.15115}
\showURL{%
\tempurl}


\bibitem[Rawles et~al\mbox{.}(2024)]%
        {rawles2024androidworld}
\bibfield{author}{\bibinfo{person}{Christopher Rawles}, \bibinfo{person}{Sarah Clinckemaillie}, \bibinfo{person}{Yifan Chang}, \bibinfo{person}{Jonathan Waltz}, \bibinfo{person}{Gabrielle Lau}, \bibinfo{person}{Marybeth Fair}, \bibinfo{person}{Alice Li}, \bibinfo{person}{William Bishop}, \bibinfo{person}{Wei Li}, \bibinfo{person}{Folawiyo Campbell-Ajala}, {et~al\mbox{.}}} \bibinfo{year}{2024}\natexlab{}.
\newblock \showarticletitle{Androidworld: A dynamic benchmarking environment for autonomous agents}.
\newblock \bibinfo{journal}{\emph{arXiv preprint arXiv:2405.14573}} (\bibinfo{year}{2024}).
\newblock


\bibitem[Rawles et~al\mbox{.}(2023)]%
        {rawlesAndroidWildLargeScale2023}
\bibfield{author}{\bibinfo{person}{Christopher Rawles}, \bibinfo{person}{Alice Li}, \bibinfo{person}{Daniel Rodriguez}, \bibinfo{person}{Oriana Riva}, {and} \bibinfo{person}{Timothy Lillicrap}.} \bibinfo{year}{2023}\natexlab{}.
\newblock \bibinfo{title}{Android in the {{Wild}}: {{A Large-Scale Dataset}} for {{Android Device Control}}}.
\newblock
\showeprint[arxiv]{2307.10088}~[cs]


\bibitem[Shinn et~al\mbox{.}(2023)]%
        {shinn2023reflexion}
\bibfield{author}{\bibinfo{person}{Noah Shinn}, \bibinfo{person}{Federico Cassano}, \bibinfo{person}{Ashwin Gopinath}, \bibinfo{person}{Karthik Narasimhan}, {and} \bibinfo{person}{Shunyu Yao}.} \bibinfo{year}{2023}\natexlab{}.
\newblock \showarticletitle{Reflexion: Language agents with verbal reinforcement learning}.
\newblock \bibinfo{journal}{\emph{Advances in Neural Information Processing Systems}}  \bibinfo{volume}{36} (\bibinfo{year}{2023}), \bibinfo{pages}{8634--8652}.
\newblock


\bibitem[Shvo et~al\mbox{.}(2021)]%
        {shvoAppBuddyLearningAccomplish2021}
\bibfield{author}{\bibinfo{person}{Maayan Shvo}, \bibinfo{person}{Zhiming Hu}, \bibinfo{person}{Rodrigo~Toro Icarte}, \bibinfo{person}{Iqbal Mohomed}, \bibinfo{person}{Allan Jepson}, {and} \bibinfo{person}{Sheila~A. McIlraith}.} \bibinfo{year}{2021}\natexlab{}.
\newblock \bibinfo{title}{{{AppBuddy}}: {{Learning}} to {{Accomplish Tasks}} in {{Mobile Apps}} via {{Reinforcement Learning}}}.
\newblock
\showeprint[arxiv]{2106.00133}~[cs]


\bibitem[Song et~al\mbox{.}(2025)]%
        {song2025mindgapexaminingselfimprovement}
\bibfield{author}{\bibinfo{person}{Yuda Song}, \bibinfo{person}{Hanlin Zhang}, \bibinfo{person}{Carson Eisenach}, \bibinfo{person}{Sham Kakade}, \bibinfo{person}{Dean Foster}, {and} \bibinfo{person}{Udaya Ghai}.} \bibinfo{year}{2025}\natexlab{}.
\newblock \bibinfo{title}{Mind the Gap: Examining the Self-Improvement Capabilities of Large Language Models}.
\newblock
\showeprint[arxiv]{2412.02674}~[cs.CL]
\urldef\tempurl%
\url{https://arxiv.org/abs/2412.02674}
\showURL{%
\tempurl}


\bibitem[Wang et~al\mbox{.}(2024d)]%
        {wang2024mobile}
\bibfield{author}{\bibinfo{person}{Junyang Wang}, \bibinfo{person}{Haiyang Xu}, \bibinfo{person}{Jiabo Ye}, \bibinfo{person}{Ming Yan}, \bibinfo{person}{Weizhou Shen}, \bibinfo{person}{Ji Zhang}, \bibinfo{person}{Fei Huang}, {and} \bibinfo{person}{Jitao Sang}.} \bibinfo{year}{2024}\natexlab{d}.
\newblock \showarticletitle{Mobile-Agent: Autonomous Multi-Modal Mobile Device Agent with Visual Perception}.
\newblock \bibinfo{journal}{\emph{arXiv preprint arXiv:2401.16158}} (\bibinfo{year}{2024}).
\newblock


\bibitem[Wang et~al\mbox{.}(2024a)]%
        {wang2024mobileagentbench}
\bibfield{author}{\bibinfo{person}{Luyuan Wang}, \bibinfo{person}{Yongyu Deng}, \bibinfo{person}{Yiwei Zha}, \bibinfo{person}{Guodong Mao}, \bibinfo{person}{Qinmin Wang}, \bibinfo{person}{Tianchen Min}, \bibinfo{person}{Wei Chen}, {and} \bibinfo{person}{Shoufa Chen}.} \bibinfo{year}{2024}\natexlab{a}.
\newblock \showarticletitle{MobileAgentBench: An Efficient and User-Friendly Benchmark for Mobile LLM Agents}.
\newblock \bibinfo{journal}{\emph{arXiv preprint arXiv:2406.08184}} (\bibinfo{year}{2024}).
\newblock


\bibitem[Wang et~al\mbox{.}(2024b)]%
        {wangMobileAgentBenchEfficientUserFriendly2024}
\bibfield{author}{\bibinfo{person}{Luyuan Wang}, \bibinfo{person}{Yongyu Deng}, \bibinfo{person}{Yiwei Zha}, \bibinfo{person}{Guodong Mao}, \bibinfo{person}{Qinmin Wang}, \bibinfo{person}{Tianchen Min}, \bibinfo{person}{Wei Chen}, {and} \bibinfo{person}{Shoufa Chen}.} \bibinfo{year}{2024}\natexlab{b}.
\newblock \bibinfo{title}{{{MobileAgentBench}}: {{An Efficient}} and {{User-Friendly Benchmark}} for {{Mobile LLM Agents}}}.
\newblock
\showeprint[arxiv]{2406.08184}~[cs]


\bibitem[Wang et~al\mbox{.}(2024c)]%
        {wang2024distrl}
\bibfield{author}{\bibinfo{person}{Taiyi Wang}, \bibinfo{person}{Zhihao Wu}, \bibinfo{person}{Jianheng Liu}, \bibinfo{person}{Jianye Hao}, \bibinfo{person}{Jun Wang}, {and} \bibinfo{person}{Kun Shao}.} \bibinfo{year}{2024}\natexlab{c}.
\newblock \showarticletitle{Distrl: An asynchronous distributed reinforcement learning framework for on-device control agents}.
\newblock \bibinfo{journal}{\emph{arXiv preprint arXiv:2410.14803}} (\bibinfo{year}{2024}).
\newblock


\bibitem[Wang et~al\mbox{.}(2024e)]%
        {wang2024ponder}
\bibfield{author}{\bibinfo{person}{Yiqin Wang}, \bibinfo{person}{Haoji Zhang}, \bibinfo{person}{Jingqi Tian}, {and} \bibinfo{person}{Yansong Tang}.} \bibinfo{year}{2024}\natexlab{e}.
\newblock \showarticletitle{Ponder \& press: Advancing visual gui agent towards general computer control}.
\newblock \bibinfo{journal}{\emph{arXiv preprint arXiv:2412.01268}} (\bibinfo{year}{2024}).
\newblock


\bibitem[Wei et~al\mbox{.}(2022)]%
        {wei2022chain}
\bibfield{author}{\bibinfo{person}{Jason Wei}, \bibinfo{person}{Xuezhi Wang}, \bibinfo{person}{Dale Schuurmans}, \bibinfo{person}{Maarten Bosma}, \bibinfo{person}{Fei Xia}, \bibinfo{person}{Ed Chi}, \bibinfo{person}{Quoc~V Le}, \bibinfo{person}{Denny Zhou}, {et~al\mbox{.}}} \bibinfo{year}{2022}\natexlab{}.
\newblock \showarticletitle{Chain-of-thought prompting elicits reasoning in large language models}.
\newblock \bibinfo{journal}{\emph{Advances in neural information processing systems}}  \bibinfo{volume}{35} (\bibinfo{year}{2022}), \bibinfo{pages}{24824--24837}.
\newblock


\bibitem[Wen et~al\mbox{.}(2024a)]%
        {wen2024autodroid}
\bibfield{author}{\bibinfo{person}{Hao Wen}, \bibinfo{person}{Yuanchun Li}, \bibinfo{person}{Guohong Liu}, \bibinfo{person}{Shanhui Zhao}, \bibinfo{person}{Tao Yu}, \bibinfo{person}{Toby Jia-Jun Li}, \bibinfo{person}{Shiqi Jiang}, \bibinfo{person}{Yunhao Liu}, \bibinfo{person}{Yaqin Zhang}, {and} \bibinfo{person}{Yunxin Liu}.} \bibinfo{year}{2024}\natexlab{a}.
\newblock \showarticletitle{Autodroid: Llm-powered task automation in android}. In \bibinfo{booktitle}{\emph{Proceedings of the 30th Annual International Conference on Mobile Computing and Networking}}. \bibinfo{pages}{543--557}.
\newblock


\bibitem[Wen et~al\mbox{.}(2024b)]%
        {wenAutoDroidV2BoostingSLMbased2024}
\bibfield{author}{\bibinfo{person}{Hao Wen}, \bibinfo{person}{Shizuo Tian}, \bibinfo{person}{Borislav Pavlov}, \bibinfo{person}{Wenjie Du}, \bibinfo{person}{Yixuan Li}, \bibinfo{person}{Ge Chang}, \bibinfo{person}{Shanhui Zhao}, \bibinfo{person}{Jiacheng Liu}, \bibinfo{person}{Yunxin Liu}, \bibinfo{person}{Ya-Qin Zhang}, {and} \bibinfo{person}{Yuanchun Li}.} \bibinfo{year}{2024}\natexlab{b}.
\newblock \bibinfo{title}{{{AutoDroid-V2}}: {{Boosting SLM-based GUI Agents}} via {{Code Generation}}}.
\newblock
\showeprint{2412.18116}~[cs]
\href{https://doi.org/10.48550/arXiv.2412.18116}{doi:\nolinkurl{10.48550/arXiv.2412.18116}}


\bibitem[Xing et~al\mbox{.}(2024)]%
        {xing2024understanding}
\bibfield{author}{\bibinfo{person}{Mingzhe Xing}, \bibinfo{person}{Rongkai Zhang}, \bibinfo{person}{Hui Xue}, \bibinfo{person}{Qi Chen}, \bibinfo{person}{Fan Yang}, {and} \bibinfo{person}{Zhen Xiao}.} \bibinfo{year}{2024}\natexlab{}.
\newblock \showarticletitle{Understanding the Weakness of Large Language Model Agents within a Complex Android Environment}.
\newblock \bibinfo{journal}{\emph{arXiv preprint arXiv:2402.06596}} (\bibinfo{year}{2024}).
\newblock


\bibitem[Xu et~al\mbox{.}(2024)]%
        {xu2024androidlab}
\bibfield{author}{\bibinfo{person}{Yifan Xu}, \bibinfo{person}{Xiao Liu}, \bibinfo{person}{Xueqiao Sun}, \bibinfo{person}{Siyi Cheng}, \bibinfo{person}{Hao Yu}, \bibinfo{person}{Hanyu Lai}, \bibinfo{person}{Shudan Zhang}, \bibinfo{person}{Dan Zhang}, \bibinfo{person}{Jie Tang}, {and} \bibinfo{person}{Yuxiao Dong}.} \bibinfo{year}{2024}\natexlab{}.
\newblock \showarticletitle{Androidlab: Training and systematic benchmarking of android autonomous agents}.
\newblock \bibinfo{journal}{\emph{arXiv preprint arXiv:2410.24024}} (\bibinfo{year}{2024}).
\newblock


\bibitem[Yang et~al\mbox{.}(2024)]%
        {yang2024aria}
\bibfield{author}{\bibinfo{person}{Yuhao Yang}, \bibinfo{person}{Yue Wang}, \bibinfo{person}{Dongxu Li}, \bibinfo{person}{Ziyang Luo}, \bibinfo{person}{Bei Chen}, \bibinfo{person}{Chao Huang}, {and} \bibinfo{person}{Junnan Li}.} \bibinfo{year}{2024}\natexlab{}.
\newblock \showarticletitle{Aria-UI: Visual Grounding for GUI Instructions}.
\newblock \bibinfo{journal}{\emph{arXiv preprint arXiv:2412.16256}} (\bibinfo{year}{2024}).
\newblock


\bibitem[Yao et~al\mbox{.}(2023)]%
        {yao2023react}
\bibfield{author}{\bibinfo{person}{Shunyu Yao}, \bibinfo{person}{Jeffrey Zhao}, \bibinfo{person}{Dian Yu}, \bibinfo{person}{Nan Du}, \bibinfo{person}{Izhak Shafran}, \bibinfo{person}{Karthik Narasimhan}, {and} \bibinfo{person}{Yuan Cao}.} \bibinfo{year}{2023}\natexlab{}.
\newblock \showarticletitle{React: Synergizing reasoning and acting in language models}. In \bibinfo{booktitle}{\emph{International Conference on Learning Representations (ICLR)}}.
\newblock


\bibitem[You et~al\mbox{.}(2024)]%
        {youFerretUIGroundedMobile2024}
\bibfield{author}{\bibinfo{person}{Keen You}, \bibinfo{person}{Haotian Zhang}, \bibinfo{person}{Eldon Schoop}, \bibinfo{person}{Floris Weers}, \bibinfo{person}{Amanda Swearngin}, \bibinfo{person}{Jeffrey Nichols}, \bibinfo{person}{Yinfei Yang}, {and} \bibinfo{person}{Zhe Gan}.} \bibinfo{year}{2024}\natexlab{}.
\newblock \bibinfo{title}{Ferret-{{UI}}: {{Grounded Mobile UI Understanding}} with {{Multimodal LLMs}}}.
\newblock
\showeprint[arxiv]{2404.05719}~[cs]


\bibitem[Zhang et~al\mbox{.}(2024)]%
        {zhangLlamaTouchFaithfulScalable2024}
\bibfield{author}{\bibinfo{person}{Li Zhang}, \bibinfo{person}{Shihe Wang}, \bibinfo{person}{Xianqing Jia}, \bibinfo{person}{Zhihan Zheng}, \bibinfo{person}{Yunhe Yan}, \bibinfo{person}{Longxi Gao}, \bibinfo{person}{Yuanchun Li}, {and} \bibinfo{person}{Mengwei Xu}.} \bibinfo{year}{2024}\natexlab{}.
\newblock \bibinfo{title}{{{LlamaTouch}}: {{A Faithful}} and {{Scalable Testbed}} for {{Mobile UI Automation Task Evaluation}}}.
\newblock
\showeprint[arxiv]{2404.16054}~[cs]


\bibitem[Zheng et~al\mbox{.}(2024)]%
        {zheng2024seeact}
\bibfield{author}{\bibinfo{person}{Boyuan Zheng}, \bibinfo{person}{Boyu Gou}, \bibinfo{person}{Jihyung Kil}, \bibinfo{person}{Huan Sun}, {and} \bibinfo{person}{Yu Su}.} \bibinfo{year}{2024}\natexlab{}.
\newblock \showarticletitle{GPT-4V(ision) is a Generalist Web Agent, if Grounded}. In \bibinfo{booktitle}{\emph{Forty-first International Conference on Machine Learning}}.
\newblock
\urldef\tempurl%
\url{https://openreview.net/forum?id=piecKJ2DlB}
\showURL{%
\tempurl}


\end{thebibliography}
\end{document}